\documentclass[letterpaper]{article} %
\usepackage[preprint]{aaai2027}  %
\usepackage[hyphens]{url}  %
\usepackage{graphicx} %
\usepackage{natbib}  %
\usepackage{caption} %
\usepackage{amsmath}
\usepackage{amssymb}
\DeclareMathOperator{\Ex}{\mathbb{E}}
\DeclareMathOperator*{\argmin}{arg\,min}

\usepackage{algorithm}
\usepackage{algorithmic}
\usepackage{booktabs}
\usepackage[table]{xcolor}
\definecolor{mycolor}{HTML}{F0F8FF} %

\title{Align, Then Correct: Training-Free Two-Stage Low-Rank Compensation for Extremely Quantized Large Language Models}
\author{
    Seobin Song\textsuperscript{\rm 1},
    Geonho Lee\textsuperscript{\rm 1},
    Janghwan Lee\textsuperscript{\rm 2}\thanks{Work done while at Hanyang University.},
    Jungwook Choi\textsuperscript{\rm 1}\corresponding
}
\affiliations{
    \textsuperscript{\rm 1}Hanyang University, Seoul, Republic of Korea\\
    \textsuperscript{\rm 2}Qualcomm AI Research, Qualcomm Korea YH, Seoul, Republic of Korea\\
    \texttt{\{skmssb, thisisho, choij\}@hanyang.ac.kr}, \texttt{janghwan@qti.qualcomm.com}
}

\begin{document}

\maketitle

\begin{abstract}
Low-rank quantization error compensation (LQEC) recovers the accuracy lost
under aggressive weight quantization by attaching a closed-form rank-$r$
adapter beside each frozen quantized weight, without any training. We show
that existing compensators are limited by two shared simplifications. They
calibrate \emph{symmetrically}, evaluating the full-precision and compensated
weights on the same activation, which yields a compensation target that is
inherently high-rank---so a fixed rank budget captures only a small fraction
of it. And they minimize only the second-order term of the loss, although the
compensated model is not stationary: a first-order descent direction larger
than the applied compensation itself remains in every layer, and no
reconstruction objective can absorb it. We propose a two-stage closed-form
framework that removes both simplifications. Stage~1 aligns each layer's
output with the full-precision model under a Fisher-weighted
\emph{asymmetric} objective, concentrating the rank budget on a
rank-compressible target. Stage~2 re-measures statistics on the compensated
model and applies a rank-constrained natural-gradient step that absorbs the
remaining first-order signal. Every adapter is the result of a single
truncated SVD; backward passes serve only to collect statistics. At $2$ bits
under QuIP\#, our method reduces WikiText-2 perplexity from $12.43$ to
$10.26$ on Qwen3-8B and from $21.11$ to $13.22$ on Qwen3-4B. On the
held-out C4 corpus, it recovers $51\%$ and $84\%$ of the gap to FP16,
versus $31\%$ and $63\%$ for the strongest baseline, with consistent gains
in the seven-task zero-shot average, at higher bit-widths, and under a
distinct quantizer.
\end{abstract}

\section{Introduction}
\label{sec:intro}

As large language models (LLMs) are increasingly deployed on memory-constrained
on-device platforms, compressing pretrained models into a smaller memory
footprint has become essential, and weight quantization is the primary means of
doing so. Post-training quantization (PTQ)
\citep{frantar2022gptq,quipsharp,brecq,li2025gptaq} is now near-lossless at $4$
bits, but at $2$ bits it still incurs severe accuracy degradation. LoRA-based
quantization error compensation (QEC) addresses this gap by attaching a
low-rank adapter $AB^\top$ beside each frozen quantized weight, absorbing the
quantization error in continuous precision at a small memory overhead. The
compensation objective has evolved steadily: QLoRA \citep{qlora} initializes
the adapters of a quantized backbone to zero, leaving compensation to
fine-tuning; LoftQ \citep{loftq} fits the best rank-$r$ approximation of the
weight error $W-\hat W$ in closed form \citep{eckartyoung}; LQ-LoRA
\citep{lqlora} adds loss sensitivity through a diagonal Fisher weighting; and
LQER \citep{lqer}, QERA \citep{qera}, EoRA \citep{eora}, CLoQ \citep{cloq}, and
ProjQ \citep{projq} target the layer output error under the full input second
moment. The appeal of this line is that it is training-free: each adapter is a
closed-form solution, computed from a small calibration set in minutes on a
single GPU and composable with any quantizer. Training-based alternatives
\citep{rilq} recover more accuracy but forfeit exactly this property, requiring
renewed training for every bit-width, rank, and model.

Despite this progress, existing training-free compensators share two
structural simplifications. First, they calibrate \emph{symmetrically}: the
full-precision and compensated weights are evaluated on the same activation,
so the adapter's target reduces to $(W-\hat W)\hat X$. This target is poorly
matched to a small rank budget---its energy spreads across many singular
directions, so a rank-$64$ adapter captures only a small fraction of it. The
\emph{asymmetric} target $WX-\hat W\hat X$, formed from the activations the
full-precision and quantized models actually produce, concentrates its
energy in far fewer directions, yet no existing objective distinguishes the
two activations. Second, these methods minimize only the quadratic term of
the loss expansion, keeping the convergence assumption that the gradient
vanishes even though quantization has displaced the model from its converged
point \citep{brecq}. In practice, the first-order descent direction that
remains after even a strong second-order compensation exceeds the applied
compensation in every layer---by more than an order of magnitude in most layers---and no
reconstruction objective can absorb it, because such objectives contain no
first-order term.

We propose a two-stage low-rank compensation method that removes each simplification
in turn, entirely in closed form. Stage~1 solves the Fisher-weighted
asymmetric objective: each adapter acts on the quantized activation it
actually receives and is solved to match the full-precision output, with the
output axis weighted by the full empirical Fisher so that the rank budget
follows loss sensitivity; the blocks are solved front to back with input
statistics refreshed at each step. Stage~2 re-measures every statistic on
the compensated model, block by block, and updates each adapter with a
single natural-gradient step
\citep{amari1998natural,martens2015optimizing}, recovering the first-order
term; carrying the existing adapter inside the target makes this a
generalized rank-constrained approximation \citep{friedland2007generalized}
whose solution has rank $r$ by construction. In this framework, every
existing training-free compensator solves a Stage-1-type reconstruction
objective under coarser statistics---none uses the full output Fisher, and
none distinguishes the two activations---whereas Stage~2, a closed-form step
taken against the task loss on the compensated model itself, has no
counterpart in prior work. Unlike asymmetric quantizer-side calibration
\citep{li2025gptaq,qep}, whose correction must ultimately be absorbed by the
quantization grid, our correction resides in continuous parameters beside
the grid and composes with any quantizer. Throughout, backward passes serve
only to collect statistics; no iterative optimization is performed.

We evaluate across models and quantizers---QuIP\# \citep{quipsharp} at
$2$--$4$ bits and groupwise round-to-nearest---with the same rank budget as
all baselines. At $2$ bits under QuIP\#, our method lowers WikiText-2
perplexity from $12.43$ to $10.26$ on Qwen3-8B and from $21.11$ to $13.22$
on Qwen3-4B. On the held-out C4 corpus, it closes $51\%$ and $84\%$ of the
gap to FP16 where the strongest baseline closes $31\%$ and $63\%$, and the
gains carry over to seven zero-shot common-sense tasks, higher bit-widths,
and the distinct quantizer.

Our contributions are summarized as follows:
\begin{itemize}
\item \textbf{Analysis.} Identification of two simplifications shared by all
training-free compensators, with direct measurements of their cost:
symmetric calibration yields a target that resists low-rank approximation,
and a dominant first-order signal survives any second-order compensation
(Sec.~\ref{sec:motivation}).
\item \textbf{Stage~1.} A closed-form Fisher-weighted asymmetric alignment,
the first LQEC objective to use the full output Fisher and to distinguish
the full-precision and quantized activations (Sec.~\ref{sec:method}).
\item \textbf{Stage~2.} To our knowledge, the first training-free
compensation step solved against the compensated model itself, recovering
the first-order loss term in closed form under the rank constraint
(Sec.~\ref{sec:method}).
\item \textbf{Results.} Best perplexity and seven-task average accuracy among
training-free compensators across bit-widths and quantizers: at $2$ bits on
Qwen3, $51$--$84\%$ of the held-out C4 gap to FP16 is recovered versus at
most $63\%$ for the strongest baseline, with Stage~2 contributing the largest gain using
only $8$K calibration tokens (Sec.~\ref{sec:experiments}).
\end{itemize}

\section{Background}
\providecommand{\Wq}{\hat{W}}
\providecommand{\xq}{\hat{x}}
\providecommand{\Xq}{\hat{X}}
\providecommand{\yq}{\hat{y}}
\providecommand{\Hx}{H_{\xq}}
\providecommand{\Hy}{H_{\yq}}
\providecommand{\Eres}{E_{\mathrm{res}}}
\providecommand{\Edr}{E_{\mathrm{dr}}}
\providecommand{\diag}{\operatorname{diag}}
\providecommand{\Frow}{D_{\mathrm{row}}}
\providecommand{\Fcol}{D_{\mathrm{col}}}
\label{sec2}

\subsection{Second-Order Analysis of Quantization Error}
\label{subsec:related-second}

To reduce the accuracy degradation of a quantized model, a principled approach
is to quantify how the weight perturbation $\Delta w$ introduced by
quantization changes the expected loss $\mathcal{L}$. Writing
$w=\mathrm{vec}(W)$ for the vectorized weights of a linear layer, a
second-order Taylor expansion of the loss degradation $\Delta\mathcal{L}$
around the pretrained model gives:
\begin{equation}
\mathbb{E}[\mathcal{L}(w+\Delta w)]-\mathbb{E}[\mathcal{L}(w)]
\approx
\bar{g}^{\top}\Delta w
+
\frac{1}{2}\Delta w^{\top}\bar{H}\Delta w,
\label{eq:taylor}
\end{equation}
where $\bar{g}=\mathbb{E}[\nabla_w\mathcal{L}]$ and
$\bar{H}=\mathbb{E}[\nabla_w^2\mathcal{L}]$ are the expected gradient and
Hessian. Existing quantization methods assume the pretrained model has
converged, so that $\bar{g}\approx 0$, and retain only the quadratic term.

The weight-space Hessian $\bar{H}$ is too large to compute at LLM scale. For a
linear layer, the quadratic term is therefore approximated in two steps:
\begin{equation}
\Delta\mathcal{L}
\approx
\frac{1}{2}\mathbb{E}_t\!\left[
\Delta z_t^{\top}H_{z,t}\Delta z_t
\right]
\approx
\frac{1}{2}\mathbb{E}_t\!\left[
\Delta z_t^{\top}H_z\Delta z_t
\right],
\label{eq:zHz-fisher}
\end{equation}
where $t$ indexes calibration tokens. The first step rewrites the quadratic
term in the layer's output space~\cite{brecq}: $\Delta z_t=z_t-\hat{z}_t$ is
the difference between the outputs of the full-precision and quantized linear
layers, and $H_{z,t}$ is the loss Hessian with respect to the output $z_t$,
which is small enough to compute but varies with every token. The second step
therefore replaces $H_{z,t}$ with the calibration-averaged empirical Fisher
$H_z=\mathbb{E}_t[g_tg_t^{\top}]$, where
$g_t=\nabla_{z_t}\mathcal{L}_t$~\cite{singh2020woodfisher,brecq}; $H_z$ is positive
semidefinite by construction and requires only per-token output gradients from
a backward pass over the calibration set. Unless otherwise specified, $H_z$
denotes this matrix throughout the paper. Eq.~\eqref{eq:zHz-fisher} quantifies
the quantization-induced loss degradation that PTQ techniques aim to
minimize~\cite{frantar2022gptq,brecq,quipsharp}.

At a high level, the curvature approximation used here follows three steps.
The intractable weight Hessian is first replaced by the weight Fisher, which is
then estimated on calibration tokens by the empirical Fisher. For a linear
layer, K-FAC~\cite{martens2015optimizing} further approximates this empirical
weight Fisher by a Kronecker product,
$F_W^{\mathrm{emp}}\approx H_x\otimes H_z$, where $H_x$ is the input second
moment and the output factor $H_z$ is the empirical Fisher defined above.
Thus, the factors used throughout this paper arise from the sequence
Hessian $\rightarrow$ Fisher $\rightarrow$ empirical Fisher $\rightarrow$
K-FAC.

\subsection{Low-Rank Quantization Error Compensation}
\label{subsec:related-lqec}

Even state-of-the-art PTQ methods incur severe accuracy degradation under
aggressive quantization at around $2$ bits. Low-rank quantization error
compensation (LQEC) addresses this gap by attaching a rank-$r$ adapter
$AB^{\top}$ beside each frozen quantized weight $\Wq$, absorbing part of the
quantization error in continuous precision at a small memory overhead. With
the adapter attached, the output perturbation becomes:
\begin{equation}
\Delta z
=
Wx-
(\Wq+AB^{\top})\hat{x},
\label{eq:lqec-dz}
\end{equation}
where $x$ and $\hat{x}$ denote the inputs this layer receives in the
full-precision and quantized models, respectively. Existing LQEC methods
obtain $A$ and $B$ in closed form from calibration statistics; they differ
only in which approximations of Eq.~\eqref{eq:taylor} they solve.

\paragraph{Common assumptions.}
All existing methods share two simplifications. First, they calibrate
activations \textit{symmetrically}: the full-precision and compensated
weights are evaluated on a shared input $\bar{x}\in\{x,\hat{x}\}$, collapsing
the perturbation to $\Delta z=(W-\Wq-AB^{\top})\bar{x}$. The discrepancy
between $x$ and $\hat{x}$, discussed on the quantizer side by
GPTAQ~\cite{li2025gptaq}, has not been addressed in any LQEC objective.
Second, they minimize only the quadratic term of Eq.~\eqref{eq:taylor},
keeping the convergence assumption $\bar{g}\approx 0$ even though
quantization has already moved the model away from its converged point.

\paragraph{Differences in approximation.}
Within this shared form, the methods are distinguished by how they treat the
output-side empirical Fisher $H_z$ and the input second moment via K-FAC
approximation~\cite{martens2015optimizing}. %
LoftQ~\cite{loftq} sets both to identity, reducing the problem to a truncated
SVD of the weight error $W-\Wq$.
LQER~\cite{lqer}, QERA~\cite{qera}, EoRA~\cite{eora}, and CLoQ~\cite{cloq}
keep $H_z=I$ but weight the input axis with the full activation second
moment, yielding a whitened SVD; ProjQ~\cite{projq} additionally re-optimizes
$\Wq$ jointly with the adapter.
LQ-LoRA~\cite{lqlora} instead introduces loss sensitivity, approximating
$H_z$ by its \textit{diagonal} through a rank-one factorization of the
empirical Fisher. No existing method uses the full $H_z$. We revisit the two
shared simplifications above in Sec.~\ref{sec:motivation}.

\begin{figure}[t]
\centering
\includegraphics[
    width=\columnwidth,
    trim=339 205 365 164,
    clip
]{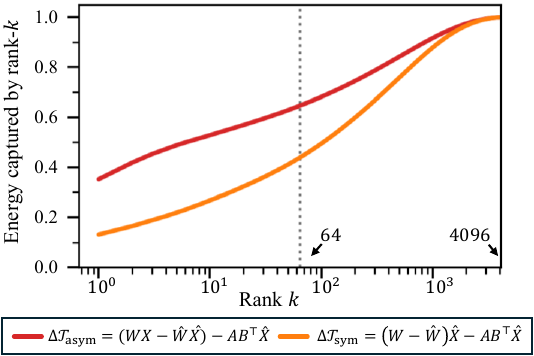}
\caption{Captured energy of the rank-$k$ truncation,
$\big(\sum_{i=1}^{k}\sigma_i^2\big)\big/\big(\sum_{i=1}^{d}\sigma_i^2\big)$,
for the symmetric and asymmetric compensation objectives, where $\sigma_i$
denotes the $i$-th singular value and $d=\min(m,n)$. At $k=64$, the
asymmetric objective captures substantially more energy than the symmetric
one, indicating that a fixed rank budget discards less information under the
asymmetric objective. The asymmetric objective is therefore inherently more
compatible with low-rank compensation. Computed on \texttt{mlp.down\_proj}
of layer 0 in Qwen3-8B quantized to 2 bits, calibrated on
$8\times2048=16{,}384$ WikiText-2 tokens.}
\label{fig:obj-rank}
\end{figure}

\begin{figure}[t]
\centering

\includegraphics[width=\columnwidth]{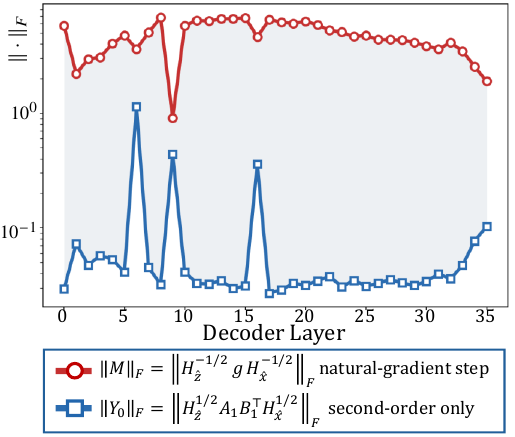}
\caption{Layer-wise whitened Frobenius norms of the Stage-1 adapter,
$\|Y_0\|_F=\|H_{\hat z}^{1/2}A_1B_1^\top H_{\hat x}^{1/2}\|_F$ (blue; the
second-order compensation), and of the natural-gradient step on the
compensated model, $\|M\|_F=\|H_{\hat z}^{-1/2}gH_{\hat x}^{-1/2}\|_F$ (red).
Both are expressed in the whitened coordinates induced by the curvature
metric of Eq.~\eqref{eq:zHz-fisher}, so their magnitudes are directly
comparable; each point averages the seven linear modules of a decoder layer,
and the vertical axis is logarithmic. The natural-gradient step is larger in
every layer, indicating that the second-order objective alone leaves
substantial loss-sensitive information unexploited. Measured on Qwen3-4B
quantized to $2$ bits with QuIP\#, using $16\times512=8{,}192$ WikiText-2
training tokens and one forward and backward pass per block.}
\label{fig:residual-gradient}
\end{figure}

\section{Motivations}
\label{sec:motivation}

Sec.~\ref{sec2} showed that existing LQEC methods share two simplifications:
symmetric calibration and the omission of the first-order term. This section
examines each of them in the regime where LQEC is needed most---recovering
accuracy under aggressive $2$-bit quantization---and shows that neither is
justified there. First, under symmetric calibration the compensation target
is inherently high-rank, so a rank-$r$ adapter captures only a small fraction
of it. Second, after strong compensation the model no longer sits at a
stationary point, so the neglected first-order term still provides a descent
direction for the task loss.

\subsection{High-Rank Targets under Symmetric Calibration}

Because LQEC deploys a fixed-rank adapter, its attainable quality is bounded
by how much of the compensation target a rank-$r$ matrix can represent.
Written over $T$ calibration tokens, the perturbation of
Eq.~\eqref{eq:lqec-dz} splits into an adapter-independent target and an
adapter term,
$\Delta Z = (WX-\Wq\hat{X}) - AB^{\top}\hat{X}$, and the target the rank-$r$
adapter must approximate takes one of two forms depending on the calibration:
\begin{equation}
\mathcal{T}_{\mathrm{sym}} = (W-\Wq)\hat{X},
\qquad
\mathcal{T}_{\mathrm{asym}} = WX-\Wq\hat{X},
\label{eq:rank-targets}
\end{equation}
where $X=[x_1,\dots,x_T]$ and $\hat{X}=[\hat{x}_1,\dots,\hat{x}_T]$ collect
the per-token activations of Eq.~\eqref{eq:lqec-dz}.
$\mathcal{T}_{\mathrm{sym}}$ is the target prior LQEC methods adopt under the
symmetric calibration $x\approx\hat{x}$ of Sec.~\ref{subsec:related-lqec},
whereas $\mathcal{T}_{\mathrm{asym}}$ retains both activations and thus keeps
the portion of the error that the symmetric target discards.
$\Delta\mathcal{T}_{\bullet}=\mathcal{T}_{\bullet}-AB^{\top}\hat{X}$ is
accordingly the residual each objective leaves for the rank-$r$ adapter to
still account for, and its singular spectrum determines how much of that
residual a given rank can capture. Fig.~\ref{fig:obj-rank} compares this
spectrum for the two targets: at the operating rank $r=64$, the asymmetric
target concentrates substantially more of its energy in the leading
directions than the symmetric one, so a rank-$r$ adapter fit to
$\mathcal{T}_{\mathrm{asym}}$ discards less information than one fit to
$\mathcal{T}_{\mathrm{sym}}$, making the asymmetric objective inherently more
compatible with low-rank compensation. Compressibility alone does not
guarantee accuracy, however: which of the retained directions matter is
decided by the loss weighting, and the asymmetric target pays off only when
combined with the full output Fisher (Sec.~\ref{sec:experiments}).

\subsection{Remaining First-Order Signal after Compensation}

The second simplification of Sec.~\ref{subsec:related-lqec} is to drop the
first-order term of Eq.~\eqref{eq:taylor} under the convergence assumption
$\bar{g}\approx 0$. Quantization, however, displaces the model from its
converged point, and compensation restores it only partially. We therefore
ask: after second-order compensation is applied, how large is the gradient
that remains?

Fig.~\ref{fig:residual-gradient} answers this by comparing two quantities in
each layer: the low-rank compensation obtained from the second-order
objective, and the descent step that the remaining gradient prescribes on the
compensated model. Both are measured under the curvature metric induced by
Eq.~\eqref{eq:zHz-fisher}, so their norms are directly comparable in
loss-sensitive units. The remaining first-order step is larger in every
layer---by more than an order of magnitude in most layers---showing that the convergence
assumption fails on the compensated model. Moreover, no reconstruction
objective can absorb this signal, because such objectives contain no
first-order term. This untapped signal motivates an explicit first-order
correction stage.

\begin{figure*}[t]
\centering
\includegraphics[width=0.9\textwidth]{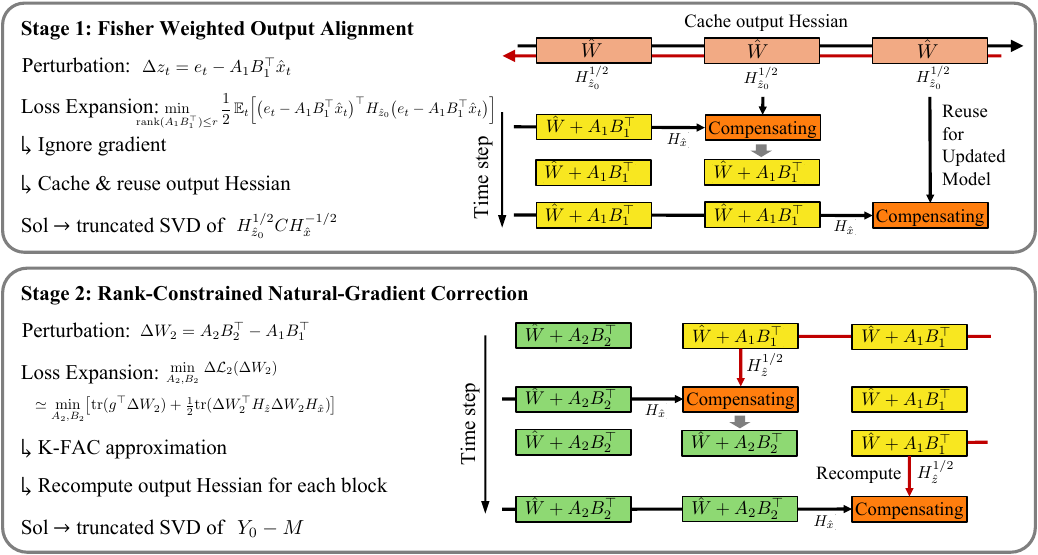}
\caption{The proposed two-stage procedure on a three-block model. Every linear layer
keeps its quantized weight $\hat W$ frozen and attaches a rank-$r$ adapter
$AB^\top$. \textbf{(a)} Stage~1 sweeps the decoder front to back. The output
Fisher $H_{\hat z_0}$ is computed once and reused, whereas the input moments
$H_{\hat x}$ and $\tilde H_x$ are refreshed before each block.
\textbf{(b)} Stage~2 repeats the sweep, collecting $g$, $H_{\hat x}$, and
$H_{\hat z}$ with one forward and backward pass per block before applying the
natural-gradient correction. Red outlines indicate the adapters being updated.}
\label{fig:method}
\end{figure*}

\section{Methodology}
\label{sec:method}

Sec.~\ref{sec:motivation} identified two sources of accuracy that existing
LQEC leaves unused: the compensation target becomes far more
rank-compressible once symmetric calibration is dropped, and a substantial
first-order signal survives any second-order solve. Our framework
(Fig.~\ref{fig:method}) converts each observation into one closed-form
stage, keeping the quantized weights frozen and writing only the rank-$r$
adapters. Stage~1 aligns each layer's output with the full-precision model
under the Fisher-weighted \textit{asymmetric} objective, spending the rank budget on
the compact target $\mathcal{T}_{\mathrm{asym}}$. Stage~2 then absorbs the
\textit{first-order signal} remaining on the compensated model through a
rank-constrained natural-gradient step. Both stages reduce to a single SVD
per layer, with no iterative optimization and no training. The complete
procedure is summarized in Algorithm~\ref{alg:ours} in
Appendix~\ref{app:algorithm}.

\subsection{Stage 1: Fisher-Weighted Output Alignment}

\subsubsection{Fisher-weighted asymmetric objective.}

To account for asymmetric calibration, we optimize the output perturbation in
Eq.~\eqref{eq:lqec-dz} using the Fisher-weighted quadratic objective in
Eq.~\eqref{eq:zHz-fisher}. For a rank-$r$ adapter $A_1B_1^\top$ attached to
the frozen quantized weight $\Wq$, define the token-level compensation target
as $e_t = Wx_t-\Wq\xq_t$, the $t$-th column of $\mathcal{T}_{\mathrm{asym}}$
in Eq.~\eqref{eq:rank-targets}. Since the adapter acts on the quantized input, the
remaining output perturbation is
$\Delta z_t=e_t-A_1B_1^\top\xq_t$, yielding
\begin{equation}
\label{eq:stage1-obj}
\min_{\operatorname{rank}(A_1B_1^\top)\le r}
\frac{1}{2}\Ex_t\!\left[
\big(e_t-A_1B_1^\top\xq_t\big)^\top
H_{\hat z_0}
\big(e_t-A_1B_1^\top\xq_t\big)
\right].
\end{equation}
Here, $H_{\hat z_0}=\Ex_t[\gamma_t\gamma_t^\top]$, with
$\gamma_t=\partial\ell_t/\partial\hat z_{0,t}$, is the full output-side
empirical Fisher: it instantiates the $H_z$ of Eq.~\eqref{eq:zHz-fisher} at
the pre-adapter quantized output $\hat z_0$. We estimate
$H_{\hat z_0}$ once on the original quantized model and reuse it throughout
Stage~1.

Let
$\tilde H_x=\Ex_t[x_t\xq_t^\top]$,
$\Hx=\Ex_t[\xq_t\xq_t^\top]$, and
$C=\Ex_t[e_t\xq_t^\top]=W\tilde H_x-\Wq\Hx.$
Then, Eq.~\eqref{eq:stage1-obj} is a Fisher-weighted reduced-rank regression
problem. Its optimal rank-$r$ solution is obtained from a single truncated SVD
(Theorem~1; proof in Appendix~\ref{app:thm1}):
\begin{equation}
\label{eq:stage1-svd}
U\Sigma V^\top
=
\operatorname{SVD}_r\!\left(
H_{\hat z_0}^{1/2}C\Hx^{-1/2}
\right).
\end{equation}
The adapter factors are recovered as:
\begin{equation}
\label{eq:stage1-factors}
A_1
=
H_{\hat z_0}^{-1/2}U\Sigma^{1/2},
\qquad
B_1^\top
=
\big(\Hx^{-1/2}V\Sigma^{1/2}\big)^\top.
\end{equation}
The left whitening by $H_{\hat z_0}^{1/2}$ ranks output-error directions by
their contribution to the loss rather than by their Euclidean magnitude.
Setting $H_{\hat z_0}=I$ reduces Eq.~\eqref{eq:stage1-obj} to unweighted
output reconstruction and the decomposition in Eq.~\eqref{eq:stage1-svd} to
an SVD of $C\Hx^{-1/2}$.

\subsubsection{Shrinking the closed-form solution.}

Because the moments in Eq.~\eqref{eq:stage1-svd} are estimated from a finite
calibration set, the retained singular values can overestimate their
population counterparts. We therefore apply only a fraction $\eta$ of the
closed-form update by replacing $A_1\leftarrow\eta A_1$ after
Eq.~\eqref{eq:stage1-factors}. All subsequent occurrences of
$A_1B_1^\top$, including those in Stage~2, refer to this shrunk adapter. This
uniform shrinkage follows the principle that singular values selected from
noisy observations are systematically inflated
\citep{gavish2017optimal}. An out-of-sample diagnostic confirms this
inflation directly: singular values re-evaluated on held-out calibration data
fall consistently below their in-sample counterparts, and the gap narrows as
the calibration set grows (Appendix~\ref{app:svshrink}).

\subsubsection{Sequential block-wise calibration.}

We solve decoder blocks from front to back and remeasure the input moments
$\tilde H_x$ and $\Hx$ before each block. All linear layers within the same
block are solved from the same measurements. Because compensating an upstream
block changes the quantized activations received downstream, refreshing these
moments prevents later adapters from fitting stale input distributions. This
procedure follows the layer-wise calibration principle of post-training
quantization \citep{hubara2021accurate}, while explicitly retaining the cross
moment $\tilde H_x=\Ex_t[x_t\xq_t^\top]$ required by the asymmetric objective.

\subsection{Stage 2: Low-Rank Natural-Gradient Correction}
\subsubsection{First-order objective.}
Stage~2 replaces the Stage-1 adapter by a new rank-$r$ adapter $A_2B_2^\top$
and optimizes their difference in weight space:
\begin{equation}
\Delta W_2=A_2B_2^{\top}-A_1B_1^{\top},
\label{eq:stage2-increment}
\end{equation}
which changes the layer output by $\Delta W_2\xq_t$. The rank constraint
applies to the final adapter $A_2B_2^\top$, not to the increment itself.

Let $\Delta\mathcal{L}_2(\Delta W_2)$ denote the loss change about the
Stage-1 effective weight $\Wq+A_1B_1^{\top}$. Because the compensated model
is not stationary (Sec.~\ref{sec:motivation}), the first-order term must now
be kept. For a linear layer, K-FAC~\citep{martens2015optimizing} factors the
weight-space empirical Fisher \citep{singh2020woodfisher} into input and
output moments:
\[
F_W=\Ex_t\!\left[(\xq_t\xq_t^\top)\otimes
(\gamma_t\gamma_t^\top)\right]
\approx H_{\hat x}\otimes H_{\hat z},
\]
where $H_{\hat x}=\Ex_t[\xq_t\xq_t^\top]$ and
$H_{\hat z}=\Ex_t[\gamma_t\gamma_t^\top]$ is the output-space empirical
Fisher. The resulting quadratic model is:
\begin{equation}
\label{eq:stage2-L}
\begin{split}
&\min_{A_2,B_2}\ \Delta\mathcal{L}_2(\Delta W_2) \\
&\overset{\mathrm{Fisher}}{\simeq}\min_{A_2,B_2}\!\left[
\mathrm{tr}(g^{\top}\Delta W_2)+\tfrac{1}{2}\Ex_t
 (\gamma_t^{\top}\Delta W_2\xq_t)^2\right] \\
&\overset{\mathrm{K\text{-}FAC}}{\simeq}\min_{A_2,B_2}\!\left[
\mathrm{tr}(g^{\top}\Delta W_2)+\tfrac{1}{2}\mathrm{tr}
(\Delta W_2^{\top}H_{\hat z}\Delta W_2H_{\hat x})\right] ,
\end{split}
\end{equation}
where $g=\Ex_t[\gamma_t\xq_t^{\top}]$ is the calibration-averaged weight
gradient, which is also the gradient with respect to $\Delta W_2$ since the
increment translates the effective weight. Intuitively, the linear term is
exactly the first-order signal of Sec.~\ref{sec:motivation} that
reconstruction objectives omit, and the quadratic term prices every weight
move by the loss curvature.

\subsubsection{Closed-form solution.}
Being quadratic, \eqref{eq:stage2-L} completes the square about its
unconstrained minimizer $-H_{\hat z}^{-1} g H_{\hat x}^{-1}$, the K-FAC
natural-gradient direction \citep{amari1998natural}. Adding that step to the
adapter already in place gives the target
$T=A_1B_1^{\top}-H_{\hat z}^{-1}gH_{\hat x}^{-1}$, and writing
$R=A_2B_2^{\top}-T$ reduces \eqref{eq:stage2-L} to:
\begin{equation}
\label{eq:stage2-obj}
\min_{A_2,B_2}\ \Delta\mathcal{L}_2(\Delta W_2) \ \simeq\
\min_{A_2,B_2}\ \tfrac{1}{2}\,
\mathrm{tr}\!\big[R^{\top} H_{\hat z}\, R\, H_{\hat x}\big] + c ,
\end{equation}
where $c$ is a constant. In words, the best rank-$r$ adapter is the one
closest---in the curvature-weighted metric---to the Stage-1 adapter moved by
one natural-gradient step. Both weighting factors are invertible, so this is
a generalized rank-constrained approximation problem
\citep{friedland2007generalized}. Define the whitened Stage-1 adapter and
natural-gradient step as
$Y_0=H_{\hat z}^{1/2}A_1B_1^\top H_{\hat x}^{1/2}$ and
$M=H_{\hat z}^{-1/2}gH_{\hat x}^{-1/2}$---the whitening realizes the
curvature metric of Sec.~\ref{sec:motivation}, under which
Fig.~\ref{fig:residual-gradient} compares these two quantities. The solution
(Theorem~2; proof in Appendix~\ref{app:thm2}) takes
\begin{equation}
U'\Sigma'V'^\top = \mathrm{SVD}(Y_0-M).
\label{eq:stage2-svd}
\end{equation}
Writing $U'_r$, $\Sigma'_r$, and $V'_r$ for the leading $r$ factors gives
\begin{equation}
\label{eq:stage2-factors}
A_2 = H_{\hat z}^{-1/2}\, U'_r\, \big(\Sigma'_r\big)^{1/2} ,
\qquad
B_2^\top = \big(H_{\hat x}^{-1/2}\, V'_r\,
\big(\Sigma'_r\big)^{1/2}\big)^\top .
\end{equation}
Truncating only the natural-gradient step $-M$ and adding it to the Stage-1
adapter could produce rank $2r$; truncating the adapter and step as a whole
keeps $A_2B_2^\top$ at rank $r$ by construction, with no re-projection.
 
\subsubsection{Trust-ratio step scaling.}
Taking the full step is not always warranted, so we scale the step in
\eqref{eq:stage2-svd} by $\alpha = \min\!\left(1,\ \tau\,
\frac{\lVert Y_0\rVert_F}{\lVert M\rVert_F}\right)$,
so that the decomposition becomes:
\begin{equation}
\label{eq:stage2-svd-alpha}
U'\Sigma'V'^{\top} = \mathrm{SVD}(Y_0-\alpha M) ,
\end{equation}
with $A_2$ and $B_2$ recovered exactly as before. The ratio $\alpha$ is the trust ratio of LARS \citep{you2017large}; the size
of the update is governed by the parameters it updates---here the Stage-1
adapter $Y_0$---rather than by the magnitude of the step $M$. The clip at
$1$ is not a tuned constant: at $\alpha=1$ the decomposed matrix is the
target $T=Y_0-M$ itself, so $A_2B_2^{\top}$ is already the exact minimizer
of \eqref{eq:stage2-obj}, and nothing in the quadratic model justifies
stepping past it. A sweep over $\tau$ supports this design: re-truncation
alone ($\tau=0$) reproduces the Stage-1 perplexity exactly, a broad stable
optimum appears at $\tau=0.2$--$0.3$, and the full step ($\tau=1$) leaves
the regime where the local quadratic model is accurate
(Appendix~\ref{app:tausweep}).

\subsubsection{Block-wise re-measurement.}
The statistics $g$, $H_{\hat z}$, and $H_{\hat x}$ are measured on the
current model, with the Stage-1 adapters merged, and are collected afresh at
every block by one forward and one backward pass. Unlike in Stage~1, the
output statistic cannot be collected once and reused: updating one block
changes not only the inputs the downstream blocks receive but also the
gradients that reach them, so both sides must be re-measured together.

\section{Experiments}
\label{sec:experiments}

\begin{table}[t]
\centering
\small
\setlength{\tabcolsep}{5pt}
\resizebox{\columnwidth}{!}{%
\begin{tabular}{@{}l ccc ccc@{}}
\toprule
& \multicolumn{3}{c}{Qwen3-8B}
& \multicolumn{3}{c}{Qwen3-4B} \\
\cmidrule(lr){2-4}\cmidrule(lr){5-7}
Method & WT2 & C4 & CSQA & WT2 & C4 & CSQA \\
\midrule
FP16  & 9.73 & 15.24 & 69.38 & 13.64 & 19.84 & 66.73 \\
\midrule
QuIP\# & 12.43 & 19.19 & 62.81 & 21.11 & 29.21 & 57.47 \\
+ EoRA & 12.20 & 18.98 & 63.03 & 20.92 & 27.71 & 58.69 \\
+ QERA & 12.22 & 18.97 & 63.06 & 20.91 & 27.67 & 58.63 \\
+ ProjQ & 12.15 & 18.91 & 63.07 & 21.16 & 27.81 & 58.41 \\
+ LQ-LoRA & 11.62 & 17.95 & 64.29 & 16.75 & 23.34 & 59.66 \\
+ Ours 
& \textbf{10.26} & \textbf{17.17} & \textbf{64.51}
& \textbf{13.22} & \textbf{21.36} & \textbf{60.05} \\
\bottomrule
\end{tabular}%
}
\caption{Main results at $2$ bits: WikiText-2 and C4 perplexity ($\downarrow$)
and the CSQA Avg, the mean of the seven zero-shot common-sense QA accuracies
($\uparrow$), under QuIP\# quantization. Calibration uses WikiText-2 only, so
C4 is held out. Best per model in \textbf{bold}; the FP16 row is a reference
and is excluded. Qwen3 uses a different tokenizer, so its perplexity should be
read against its own family. Per-model tables covering all three bit-widths and
the per-task breakdowns are given in Appendix~\ref{app:full-results}.}
\label{tab:main2bit}
\end{table}

\begin{table}[t]
\centering
\small
\setlength{\tabcolsep}{5pt}
\resizebox{\columnwidth}{!}{%
\begin{tabular}{@{}l ccc ccc ccc@{}}
\toprule
& \multicolumn{3}{c}{$3$ bits}
& \multicolumn{3}{c}{$4$ bits} \\
\cmidrule(lr){2-4}\cmidrule(lr){5-7}
Method & WT2 & C4 & CSQA & WT2 & C4 & CSQA  \\
\midrule
FP16 & 9.73 & 15.24 & 69.38 & 9.73 & 15.24 & 69.38 \\
\midrule
QuIP\# & 10.48 & 16.30 & 68.10 & 9.99 & 15.54 & 68.83 \\
+ EoRA & 10.37 & 16.22 & 68.54 & 9.94 & 15.53 & 68.76 \\
+ QERA & 10.38 & 16.21 & 68.55 & 9.95 & 15.53 & 68.91 \\
+ ProjQ & 10.36 & 16.21 & 68.49 & 9.96 & 15.55 & 68.84 \\
+ LQ-LoRA & 10.24 & 15.90 & 68.22 & 9.87 & 15.46 & 69.00 \\
+ Ours 
& \textbf{8.54} & \textbf{14.56} & \textbf{69.11}
& \textbf{8.22} & \textbf{14.15} & \textbf{69.41} \\
\bottomrule
\end{tabular}%
}
\caption{Qwen3-8B across bit-widths: WikiText-2 and C4 perplexity
($\downarrow$) and CSQA Avg ($\uparrow$). The gain persists at every bit-width
and on both corpora, including the held-out C4. Best per bit-width in
\textbf{bold}; the FP16 row is a reference and is excluded.}
\label{tab:qwen8bits}
\end{table}

\begin{table}[t]
\centering
\scriptsize
\setlength{\tabcolsep}{2.8pt}
\resizebox{\columnwidth}{!}{%
\begin{tabular}{@{}cccccccc@{}}
\toprule
Method & RTN & +EoRA & +QERA & +ProjQ & +LQ-LoRA & +Ours \\
\midrule
avg. CSQA & 56.26 & 62.26 & 62.29 & 62.16 & 61.45 & \textbf{62.50} \\
\bottomrule
\end{tabular}%
}
\caption{Average zero-shot accuracy over the seven CSQA tasks ($\uparrow$) for
LLaMA-3.1-8B under $3$-bit groupwise RTN quantization (group size $128$), with
$\eta=0.3$ and $\tau=0.1$. Best in \textbf{bold}.}
\label{tab:rtn-g128-3bit}
\end{table}

\begin{table}[t]
\small
\centering
\setlength{\tabcolsep}{6pt}
\begin{tabular}{@{}cc rrr@{}}
\toprule
Use $H_{\hat{z}_0}^{1/2}$ & Objective &  WT2\,$\downarrow$ & C4\,$\downarrow$ & CSQA Avg\,$\uparrow$ \\
\midrule
- & Sym  & 12.34 & 19.13 & 62.79 \\
- & Asym & 12.50 & 19.33 & 63.22 \\
\checkmark & Sym  & 12.26 & 19.04 & 62.92 \\
\textbf{\checkmark} & \textbf{Asym} & \textbf{10.26} & \textbf{17.17} & \textbf{64.51} \\
\bottomrule
\end{tabular}
\captionof{table}{Stage-1 ablation on Qwen3-8B at $2$ bits with QuIP\#.
Without the output Fisher $H_{\hat z_0}$, the asymmetric objective does not
improve perplexity over the symmetric one; combined with the full output
Fisher, it is best on all three metrics. The last row is the configuration
used throughout the paper, i.e., the full two-stage method (Stage~1 followed by
Stage~2), and therefore matches the last row of Table~\ref{tab:stage-order}.}
\label{tab:stage1_ablation}
\end{table}

\begin{table}[t]
\centering
\caption{Stage Ablation on Qwen3-8B at 2 bits, rank $64$, $128 \times 2048$
calibration tokens. Perplexity on WikiText-2 and on the held-out C4 (lower is
better); CSQA is the average of seven tasks (higher is better). The last row is
the configuration used everywhere else in the paper.}
\label{tab:stage-order}
\begin{tabular}{lccc}
\toprule
Configuration & WT2 $\downarrow$ & C4 $\downarrow$ & CSQA Avg $\uparrow$ \\
\midrule
S1-only                      & 12.48 & 19.33 & 63.28 \\
S2-only                      & 12.81 & 19.87 & 62.67 \\
S2 $\to$ S1                  & 12.55 & 19.31 & 63.21 \\
\textbf{S1 $\to$ S2 (ours)}  & \textbf{10.26} & \textbf{17.17} & \textbf{64.51} \\
\bottomrule
\end{tabular}
\end{table}

\subsection{Experimental Setup}

\subsubsection{Models and tasks.}
We evaluate Qwen3-8B and Qwen3-4B \citep{qwen3}, as well as
LLaMA-3.1-8B \citep{llama3} and LLaMA-3.2-1B \citep{llama32}, and report perplexity at sequence length $2048$
on two corpora: WikiText-2 \citep{wikitext2} (``WT2'') and C4
\citep{raffel2020exploring}. We further report zero-shot accuracy on seven
common-sense tasks: ARC-Challenge and ARC-Easy \citep{arc}, BoolQ
\citep{boolq}, HellaSwag \citep{hellaswag}, OpenBookQA \citep{openbookqa}, PIQA
\citep{piqa}, and WinoGrande \citep{winogrande}, scored with the LM Evaluation
Harness \citep{lmeval} (\texttt{acc\_norm} for ARC-C/E, HellaSwag, OpenBookQA,
and PIQA; \texttt{acc} for BoolQ and WinoGrande). ``CSQA'' is their
unweighted mean.

\subsubsection{Quantization.}
We use QuIP\# \citep{quipsharp} at $2$, $3$ and $4$ bits, sourced from the
official repository.\footnote{\url{https://github.com/Cornell-RelaxML/quip-sharp}}
We apply incoherence processing with the randomized Hadamard transform and
LDLQ, use codebook E8P12 at $2$ bits and E8P12RVQ3B/4B at $3$/$4$ bits, and
perform no layerwise fine-tuning. We additionally evaluate groupwise
round-to-nearest (RTN) quantization at $3$ bits with group size $128$. For this
RTN setting, the baselines and Stage~1 use $128\times2048=262$K WikiText-2
calibration tokens, while Stage~2 uses $16\times512=8$K tokens, and we set
$\eta=0.3$ and $\tau=0.1$ for this RTN setting only. All QuIP\# experiments
instead use the values in Table~\ref{tab:hparam} (Appendix~\ref{app:damping}).

\subsubsection{Baselines.}
We compare against EoRA \citep{eora}, QERA \citep{qera} (its QERA-Approx
variant), ProjQ \citep{projq}, and LQ-LoRA \citep{lqlora} with Fisher information, all at the same
uniform rank budget of $64$ per linear layer and all calibrated on
$128 \times 2048 = 262$K WikiText-2 tokens. From ProjQ we take only the low-rank
compensation stage, so that all methods share one quantizer.

\subsubsection{Our method.}
Every linear layer carries a rank-$64$ adapter, matching the rank budget given
to the baselines. Both stages draw calibration data from the WikiText-2
training split, the same source the baselines use. For both Qwen3-8B and
Qwen3-4B, Stage~1 uses $128\times2048=262$K calibration tokens, while Stage~2
uses $16\times512=8$K tokens. Every solve and evaluation runs on a single
NVIDIA A100-SXM4 GPU ($40$\,GB), and the remaining hyperparameters are listed in
the appendix.

\subsection{Main Results}

\paragraph{Performance across quantizers and bit-widths.}
Tables~\ref{tab:main2bit}--\ref{tab:rtn-g128-3bit} show that our method
attains the best perplexity and CSQA average in every setting; full
per-model and per-task results, including LLaMA-3.2-1B, are given in
Appendix~\ref{app:full-results}.

Table~\ref{tab:main2bit} reports the $2$-bit QuIP\# setting, where
compensation is needed most. Our method reduces WikiText-2 perplexity from
$12.43$ to $10.26$ on Qwen3-8B, closing about $80\%$ of the gap to the FP16
reference versus $30\%$ for the strongest baseline, and from $21.11$ to
$13.22$ on Qwen3-4B, below the FP16 reference of $13.64$. Because WikiText-2
is also the calibration corpus, we take the held-out C4 corpus as the
primary measure: perplexity drops to $17.17$ and $21.36$ (versus $17.95$ and
$23.34$ for the strongest baseline), closing $51\%$ and $84\%$ of the gap to
FP16 versus $31\%$ and $63\%$. The CSQA average likewise rises from $62.81$
to $64.51$ on Qwen3-8B and from $57.47$ to $60.05$ on Qwen3-4B; on Qwen3-4B
a clear gap to FP16 ($66.73$) remains on both C4 and CSQA.

Table~\ref{tab:qwen8bits} extends the comparison to $3$ and $4$ bits on
Qwen3-8B. The compensated model reaches lower perplexity than the FP16
reference ($8.54$ and $8.22$ versus $9.73$). This is not only an artifact
of fitting the calibration corpus: the held-out C4 perplexity also falls
below FP16 ($14.56$ and $14.15$ versus $15.24$), and the CSQA average
reaches $69.11$ and $69.41$, within $0.3$ of the FP16 reference ($69.38$)
and above every baseline, indicating that the Stage-2 gradient correction
acts as a mild, generalizing adaptation on top of error compensation.

Table~\ref{tab:rtn-g128-3bit} changes the quantizer to $3$-bit groupwise
RTN on LLaMA-3.1-8B. Compensation lifts the CSQA average from $56.26$ to
$62.50$, the highest among all methods. Because the correction resides in the adapter
rather than in the quantization grid, the same procedure applies to either
quantizer unchanged.

\paragraph{Where the gains come from.}
The ordering of the baselines at $2$ bits mirrors the taxonomy of
Sec.~\ref{sec2}. EoRA, QERA, and ProjQ, which all solve the symmetric
objective under an identity output metric, land within $0.07$ perplexity of
one another---consistent with the view that they are near-optimal solutions
of the same objective, so that further recovery requires changing the
objective rather than solving it better. LQ-LoRA, which weights the output
axis by a diagonal Fisher, improves to $11.62$. Removing the two shared
simplifications identified in Sec.~\ref{sec:motivation}---symmetric
calibration and the omission of the first-order term---accounts for the
remaining improvement to $10.26$.

\paragraph{Ablations.}
Table~\ref{tab:stage1_ablation} dissects Stage~1. The asymmetric target by
itself does not lower perplexity ($12.50$ versus $12.34$ for the symmetric
one): the additional error it exposes becomes useful only when the full
output Fisher directs the rank budget toward loss-sensitive directions,
in which case the combination reaches $10.26$. This matches
Sec.~\ref{sec:motivation}: the value of $\mathcal{T}_{\mathrm{asym}}$ lies
in how its energy is distributed, which an unweighted objective cannot
exploit. Table~\ref{tab:stage-order} shows that the two stages are
complementary rather than additive: Stage~1 alone ($12.48$), Stage~2 alone
($12.81$), and the reversed order ($12.55$) each recover little, whereas
Stage~1 followed by Stage~2 reaches $10.26$. Stage~1's role is thus not to
minimize perplexity on its own but to supply the aligned initialization from
which the natural-gradient step of Stage~2 is effective---the behavior
anticipated by the first-order analysis of Sec.~\ref{sec:motivation}.

\paragraph{Cost and data efficiency.}
Every adapter is the value of a closed-form expression, so the full
procedure involves no iterative optimization and runs on a single A100
($40$\,GB); construction times and memory are reported in
Appendix~\ref{app:cost}. Notably, Stage~2---which contributes the largest
share of the gain---uses only $8$K calibration tokens, $1/32$ of the $262$K
consumed by Stage~1 and by every baseline LQEC technique, indicating that the gradient
statistics it collects are substantially more sample-efficient than
reconstruction statistics.

\section{Conclusion}

We present a training-free, closed-form framework for low-rank quantization
error compensation via truncated SVD. Stage 1 applies Fisher-weighted
asymmetric alignment to capture rank-compressible targets, while Stage 2
leverages a rank-constrained natural-gradient step to recover remaining
first-order loss signals. At 2 bits, our method recovers $51$--$84\%$ of the
held-out C4 perplexity gap to FP16 on the Qwen3 models, versus at most $63\%$
for the strongest baseline, with consistent gains across bit-widths and quantizers.
Our framework still relies on layer-wise K-FAC curvature approximations and
a fixed quantized backbone, and Stage 2's per-block forward and backward
passes add construction cost; jointly optimizing the quantizer and the
compensator is a natural next step.

\bibliography{references}

\clearpage
\appendix
\section{Notation and Preliminaries}

\begin{center}
\scriptsize
\setlength{\tabcolsep}{1.5pt}
\begin{tabular}{@{}l l r l@{}}
\toprule
Symbol & Definition & Shape & Type \\
\midrule
\multicolumn{4}{@{}l}{\textit{Sizes and indices}} \\
$m$, $n$ & output, input dimension & & scalar \\
$r$ & rank budget & & scalar \\
$N$ & calibration-token count & & scalar \\
$N_b$, $b$ & number of blocks, block index & & scalar \\
$N_m$ & number of linear layers per block & & scalar \\
$t$, $i$, $j$ & token, output, input index & & scalar \\
\midrule
\multicolumn{4}{@{}l}{\textit{Weights and adapter}} \\
$w$, $\Delta w$ & weight vector, perturbation & $mn \times 1$ & vector \\
$W$, $\hat W$ & full-precision, quantized weight & $m \times n$ & matrix \\
$W_{\mathrm{eff}}$ & $\hat W+AB^\top$, current effective weight
  & $m \times n$ & matrix \\
$A$, $B$ & low-rank factors & $m \times r$, $n \times r$ & matrix \\
$A_1B_1^\top$ & adapter after Stage~1 & $m \times n$ & matrix \\
$A_2B_2^\top$ & adapter after Stage~2 & $m \times n$ & matrix \\
$\Delta W_2$ & $A_2B_2^\top - A_1B_1^\top$, Stage-2 increment
  & $m \times n$ & matrix \\
$D_{\mathrm{NG}}$ & natural-gradient direction & $m \times n$ & matrix \\
$T$ & Stage-2 target & $m \times n$ & matrix \\
$R$ & $A_2B_2^\top-T$ & $m \times n$ & matrix \\
\midrule
\multicolumn{4}{@{}l}{\textit{Loss and per-token quantities}} \\
$\mathcal L$ & $\Ex_t[\ell_t]$, average calibration loss & & scalar \\
$\Delta\mathcal L_2$ & Stage-2 loss change & & scalar \\
$L_2(\Delta W_2)$ & Stage-2 K-FAC quadratic model & & scalar \\
$c$ & term independent of $A_2,B_2$ & & scalar \\
$\ell_t$ & cross-entropy at token $t$ & & scalar \\
$\gamma_t$ & $\partial\ell_t/\partial\hat z_{0,t}$ in Stage~1;
$\partial\ell_t/\partial\hat z_t$ in Stage~2 & $m \times 1$ & vector \\
$g$ & $\partial \mathcal L/\partial W_{\mathrm{eff}}
=\Ex_t[\gamma_t\hat x_t^\top]$ & $m \times n$ & matrix \\
\midrule
\multicolumn{4}{@{}l}{\textit{Activations}} \\
$x_t$ & full-precision input & $n \times 1$ & vector \\
$\hat x_t$ & current low-precision input & $n \times 1$ & vector \\
$X$, $\hat X$ & $[x_1 \cdots x_N]$ per model & $n \times N$ & matrix \\
$z_t$ & $W x_t$, full-precision output & $m \times 1$ & vector \\
$\hat z_{0,t}$ & $\hat W \hat x_t$, no adapter & $m \times 1$ & vector \\
$\hat z_t$ & output with current adapter & $m \times 1$ & vector \\
$e_t$ & $W x_t-\hat W\hat x_t$, output error & $m \times 1$ & vector \\
$\Delta z_t$ & output perturbation & $m \times 1$ & vector \\
\midrule
\multicolumn{4}{@{}l}{\textit{Stage-1 compensation targets}} \\
$\mathcal T_{\mathrm{sym}}$ & $(W-\hat W)\hat X$ & $m \times N$ & matrix \\
$\mathcal T_{\mathrm{asym}}$ & $WX-\hat W\hat X$ & $m \times N$ & matrix \\
\midrule
\multicolumn{4}{@{}l}{\textit{Statistics}} \\
$H_{\hat x}$ & $\Ex_t[\hat x_t\hat x_t^\top]$ & $n \times n$ & matrix \\
$\tilde H_x$ & $\Ex_t[x_t\hat x_t^\top]$ & $n \times n$ & matrix \\
$H_{\hat z_0}$ & $\Ex_t[\gamma_t \gamma_t^\top]$ at $\hat z_{0,t}$, before sweep
  & $m \times m$ & matrix \\
$H_{\hat z}$ & $\Ex_t[\gamma_t \gamma_t^\top]$ at $\hat z_t$, per block
  & $m \times m$ & matrix \\
$C$ & $\Ex_t[e_t\hat x_t^\top]=W\tilde H_x-\hat W H_{\hat x}$
  & $m \times n$ & matrix \\
\midrule
\multicolumn{4}{@{}l}{\textit{Factors and whitened quantities}} \\
$U$, $\Sigma$, $V$ & factors of
$\operatorname{SVD}_r(H_{\hat z_0}^{1/2} C H_{\hat x}^{-1/2})$
& $m \times r$, $r \times r$, $n \times r$ & matrix \\
$U'$, $\Sigma'$, $V'$ & factors of $\operatorname{SVD}(Y_0-\alpha M)$
& $m \times r$, $r \times r$, $n \times r$ & matrix \\
$U'_r$, $\Sigma'_r$, $V'_r$ & leading rank-$r$ factors
& $m \times r$, $r \times r$, $n \times r$ & matrix \\
$Y_0$ & $H_{\hat z}^{1/2} A_1B_1^\top H_{\hat x}^{1/2}$ & $m \times n$ & matrix \\
$M$ & $H_{\hat z}^{-1/2} g H_{\hat x}^{-1/2}$ & $m \times n$ & matrix \\
\midrule
\multicolumn{4}{@{}l}{\textit{Constants}} \\
$\eta$ & Stage-1 shrinkage & & scalar \\
$\tau$ & Stage-2 relative step & & scalar \\
$\alpha$ & $\min(1, \tau \lVert Y_0\rVert_F / \lVert M\rVert_F)$ & & scalar \\
$\varepsilon_i$, $\varepsilon_o$ & damping coefficients & & scalar \\
\bottomrule
\end{tabular}
\captionof{table}{Symbols used throughout the paper. Hatted inputs come from
the quantized model in Stage~1 and the compensated model in Stage~2. A
subscript $0$ marks the pre-adapter state, and a tilde a cross-model statistic.}
\label{tab:notation}
\end{center}

All symbols refer to a single linear layer. All second moments are made positive
definite by the damping described in Appendix~\ref{app:damping}. The two output
statistics differ in their measurement points: $H_{\hat z_0}$ is collected
before Stage~1, whereas $H_{\hat z}$ is re-measured on the compensated model
during Stage~2. We write
$\operatorname{SVD}_r(M)=U\Sigma V^\top$ for the best rank-$r$ truncated SVD;
thus $U$, $\Sigma$, and $V$ already contain only the retained factors.

\paragraph{Lemma 1 (Eckart--Young--Mirsky
\citep{eckartyoung,mirsky}).}
For a matrix $K$, let $U\Sigma V^\top=\operatorname{SVD}_r(K)$. Then
\[
U\Sigma V^\top
\in \argmin_{\mathrm{rank}(Y) \le r} \lVert Y - K \rVert_F^2 .
\]

\paragraph{Lemma 2 (rank invariance).}
For invertible $P$ and $Q$,
\[
\mathrm{rank}(PAQ) = \mathrm{rank}(A) .
\]

\section{Proof of Theorem 1}
\label{app:thm1}

\paragraph{Theorem 1.}
Define
\[
C = W\tilde H_x - \hat W H_{\hat x} ,
\]
with $H_{\hat x}$ and $H_{\hat z_0}$ made positive definite by damping, and let
\[
U\Sigma V^\top
=\operatorname{SVD}_r\!\left(
H_{\hat z_0}^{1/2}\, C\, H_{\hat x}^{-1/2}\right) .
\]
Then, before shrinkage, a minimizer of \eqref{eq:stage1-obj} is
$A_1B_1^\top$ with
\[
\begin{aligned}
&A_1 = H_{\hat z_0}^{-1/2}\,U\,\Sigma^{1/2} , \\
&B_1 = H_{\hat x}^{-1/2}\,V\,\Sigma^{1/2} .
\end{aligned}
\]
\paragraph{Proof.}
\emph{Step 1 (expand the objective).}
By Eq.~\eqref{eq:lqec-dz} the token-level output perturbation is
$\Delta z_t=e_t-A_1B_1^\top\hat x_t$ with $e_t=Wx_t-\hat W\hat x_t$.
Dropping the positive factor $1/2$ from \eqref{eq:stage1-obj} and expanding
the quadratic form, we obtain the following expression, where the two cross
terms are equal because each is the transpose of the other scalar:
\[
\begin{aligned}
&\Ex_t\!\left[(e_t-A_1B_1^\top\hat x_t)^\top H_{\hat z_0}
(e_t-A_1B_1^\top\hat x_t)\right] \\
&=\Ex_t[(A_1B_1^\top\hat x_t)^\top
H_{\hat z_0}(A_1B_1^\top\hat x_t)] \\
&\phantom{={}}-2\,\Ex_t[e_t^\top H_{\hat z_0}A_1B_1^\top\hat x_t]
+\mathrm{const.}
\end{aligned}
\]
Both $A_1,B_1$-dependent terms are scalars, so they may be written as traces
and the expectation moved inside. For the quadratic term,
\[
\begin{aligned}
\Ex_t[(A_1B_1^\top\hat x_t)^\top H_{\hat z_0}(A_1B_1^\top\hat x_t)]
&=\Ex_t\!\left[\mathrm{tr}\big(B_1A_1^\top H_{\hat z_0}A_1B_1^\top
\hat x_t\hat x_t^\top\big)\right] \\
&=\mathrm{tr}(B_1A_1^\top H_{\hat z_0}A_1B_1^\top H_{\hat x}),
\end{aligned}
\]
For the cross term, the token-dependent outer product has expectation
\[
\begin{aligned}
\Ex_t[e_t\hat x_t^\top]
&=\Ex_t[(Wx_t-\hat W\hat x_t)\hat x_t^\top] \\
&=W\tilde H_x-\hat W H_{\hat x}=C .
\end{aligned}
\]
Therefore,
\[
\begin{aligned}
\Ex_t[e_t^\top H_{\hat z_0}A_1B_1^\top\hat x_t]
&=\Ex_t\!\left[\mathrm{tr}\big(
B_1A_1^\top H_{\hat z_0}e_t\hat x_t^\top\big)\right] \\
&=\mathrm{tr}(B_1A_1^\top H_{\hat z_0}C).
\end{aligned}
\]
\emph{Step 2 (complete the square in the whitened coordinates).}
Damping makes $H_{\hat x}$ and $H_{\hat z_0}$ positive definite, so their
symmetric square roots are invertible. Using the expansions from Step~1,
cyclicity of the trace and symmetry of the square roots give the first equality
below; completing the square gives the second:
\[
\begin{aligned}
&\Ex_t\!\left[(e_t-A_1B_1^\top\hat x_t)^\top H_{\hat z_0}
(e_t-A_1B_1^\top\hat x_t)\right] \\
&\quad=\mathrm{tr}(B_1A_1^\top H_{\hat z_0}A_1B_1^\top H_{\hat x})
-2\,\mathrm{tr}(B_1A_1^\top H_{\hat z_0}C)
+\text{const} \\
&\quad=\big\lVert H_{\hat z_0}^{1/2}A_1B_1^\top H_{\hat x}^{1/2}
-H_{\hat z_0}^{1/2}CH_{\hat x}^{-1/2}\big\rVert_F^2
+\text{const},
\end{aligned}
\]
Since the omitted terms are independent of $A_1,B_1$, completing the square
reduces Eq.~\eqref{eq:stage1-obj} to approximating
$H_{\hat z_0}^{1/2}CH_{\hat x}^{-1/2}$ in the Frobenius norm.

\emph{Step 3 (transfer the rank constraint).}
Multiplication by the invertible square roots is a bijection and preserves
rank (Lemma~2). The feasible whitened adapters are thus exactly the matrices
of rank at most $r$, and the problem becomes
\[
\min_{\operatorname{rank}(A_1B_1^\top)\le r}
\big\lVert H_{\hat z_0}^{1/2}A_1B_1^\top H_{\hat x}^{1/2}
-H_{\hat z_0}^{1/2}CH_{\hat x}^{-1/2}\big\rVert_F^2 .
\]

\emph{Step 4 (truncate and transform back).}
By Lemma~1 (Eckart--Young), a minimizer is the rank-$r$ truncated SVD, so
\[
\begin{aligned}
H_{\hat z_0}^{1/2}A_1B_1^\top H_{\hat x}^{1/2}
&=\operatorname{SVD}_r\!\left(
H_{\hat z_0}^{1/2}CH_{\hat x}^{-1/2}\right) \\
&=U\Sigma V^\top.
\end{aligned}
\]
Undoing the whitening and splitting $\Sigma=\Sigma^{1/2}\Sigma^{1/2}$,
\[
\begin{aligned}
A_1B_1^\top
&=H_{\hat z_0}^{-1/2}U\Sigma V^\top H_{\hat x}^{-1/2} \\
&=\big(H_{\hat z_0}^{-1/2}U\Sigma^{1/2}\big)
  \big(H_{\hat x}^{-1/2}V\Sigma^{1/2}\big)^\top ,
\end{aligned}
\]
using the symmetry of $H_{\hat x}^{-1/2}$. Reading off the factors gives
$A_1=H_{\hat z_0}^{-1/2}U\Sigma^{1/2}$ and
$B_1=H_{\hat x}^{-1/2}V\Sigma^{1/2}$, which is the closed form
\eqref{eq:stage1-svd}--\eqref{eq:stage1-factors} before shrinkage.
$\blacksquare$

\section{Proof of Theorem 2}
\label{app:thm2}
\paragraph{Weight gradient.}
Since $\hat z_{t,i} = \sum_j (W_{\mathrm{eff}})_{ij} \hat x_{t,j}$ for the
current effective weight $W_{\mathrm{eff}}$, the chain rule gives
\[
\begin{aligned}
\frac{\partial \mathcal L}{\partial (W_{\mathrm{eff}})_{ij}}
&= \Ex_t\!\left[
\underbrace{\frac{\partial \ell_t}{\partial \hat z_{t,i}}}_{\gamma_{t,i}}
\underbrace{\frac{\partial \hat z_{t,i}}
{\partial (W_{\mathrm{eff}})_{ij}}}_{\hat x_{t,j}}
\right] \\
&= \Ex_t[\gamma_{t,i}\, \hat x_{t,j}] .
\end{aligned}
\]
which in matrix form is
\[
g = \frac{\partial \mathcal L}{\partial W_{\mathrm{eff}}}
= \Ex_t[\gamma_t \hat x_t^\top] .
\]
Because the quantized weight is frozen and the Stage-1 adapter is held fixed
while the increment varies, the maps from the adapter and from the increment to
the effective weight are both translations, so the same $g$ serves as the
gradient with respect to all three.

\paragraph{The quadratic model.}
We approximate the change in the loss caused by the Stage-2 increment
$W_{\mathrm{eff}} \mapsto W_{\mathrm{eff}} + \Delta W_2$ by a quadratic
model. The increment acts on the output exactly as
\[
(W_{\mathrm{eff}} + \Delta W_2)\hat x_t
- W_{\mathrm{eff}} \hat x_t = \Delta W_2 \hat x_t .
\]
A second-order Taylor expansion in the layer output, evaluated at the current
effective weight, gives
\[
\begin{aligned}
\Delta\mathcal L_2(\Delta W_2) \approx{}
&\mathrm{tr}(g^\top\Delta W_2) \\
&+\frac12\Ex_t\!\left[
(\Delta W_2\hat x_t)^\top
\nabla_{\hat z_t}^2\ell_t\,
(\Delta W_2\hat x_t)\right].
\end{aligned}
\]

We next replace the exact output-space Hessian contribution by the
empirical-Fisher outer product,
\[
\nabla_{\hat z_t}^2\ell_t
\overset{\mathrm{Fisher}}{\approx}
\gamma_t\gamma_t^\top .
\]
Since
$(\Delta W_2\hat x_t)^\top\gamma_t\gamma_t^\top
(\Delta W_2\hat x_t)
=\big(\gamma_t^\top\Delta W_2\hat x_t\big)^2$, this gives
\[
\Delta\mathcal L_2(\Delta W_2)
\overset{\mathrm{Fisher}}{\approx}
\mathrm{tr}(g^\top\Delta W_2)
+\frac12\Ex_t\!\left[
\big(\gamma_t^\top\Delta W_2\hat x_t\big)^2\right].
\]
The quadratic term corresponds to the weight-space empirical Fisher
\[
F_W=\Ex_t\!\left[
(\hat x_t\hat x_t^\top)\otimes(\gamma_t\gamma_t^\top)\right],
\]
whose input and output factors remain coupled inside the expectation. K-FAC
\citep{martens2015optimizing} replaces this expected Kronecker product by the
product of the expectations:
\[
F_W
\approx
\underbrace{\Ex_t[\hat x_t\hat x_t^\top]}_{H_{\hat x}}
\otimes
\underbrace{\Ex_t[\gamma_t\gamma_t^\top]}_{H_{\hat z}} .
\]
Consequently, using column-wise vectorization,
\[
\begin{aligned}
&\frac12\Ex_t\!\left[
\big(\gamma_t^\top\Delta W_2\hat x_t\big)^2\right] \\
&=\frac12\Ex_t\!\left[\mathrm{tr}\big(
\Delta W_2^\top\gamma_t\gamma_t^\top\Delta W_2\,
\hat x_t\hat x_t^\top\big)\right] \\
&\overset{\mathrm{K\text{-}FAC}}{\approx}
\frac12\,\operatorname{vec}(\Delta W_2)^\top
\Big(
\underbrace{\Ex_t[\hat x_t\hat x_t^\top]}_{H_{\hat x}}
\otimes
\underbrace{\Ex_t[\gamma_t\gamma_t^\top]}_{H_{\hat z}}
\Big)
\operatorname{vec}(\Delta W_2) \\
&=\frac12\,\operatorname{vec}(\Delta W_2)^\top
\big(H_{\hat x}\otimes H_{\hat z}\big)
\operatorname{vec}(\Delta W_2) \\
&=\frac12\,\mathrm{tr}\big(
\Delta W_2^\top H_{\hat z}\Delta W_2H_{\hat x}\big).
\end{aligned}
\]
Combining the first- and second-order terms gives the quadratic model of
\eqref{eq:stage2-L}:
\[
\begin{aligned}
\Delta\mathcal L_2(\Delta W_2)
&\overset{\mathrm{K\text{-}FAC}}{\approx} L_2(\Delta W_2) \\
&=\mathrm{tr}(g^\top\Delta W_2) \\
&\quad+\frac12\,\mathrm{tr}\big(
\Delta W_2^\top H_{\hat z}\Delta W_2H_{\hat x}\big).
\end{aligned}
\]
The K-FAC approximation makes the model tractable without storing the full
$mn\times mn$ weight-space curvature.

\paragraph{Theorem 2.}
Let $A_1B_1^\top$ be the adapter Stage~1 produced and write
\[
\begin{aligned}
&Y_0 = H_{\hat z}^{1/2}\,A_1B_1^\top\,H_{\hat x}^{1/2} , \\
&M = H_{\hat z}^{-1/2}\,g\,H_{\hat x}^{-1/2} .
\end{aligned}
\]
For the unscaled solution $\alpha=1$, let
$U'\Sigma'V'^\top=\operatorname{SVD}(Y_0-M)$, and let
$U'_r$, $\Sigma'_r$, and $V'_r$ be its leading rank-$r$ factors. Then a
minimizer of
\[
\min_{A_2,B_2}\; L_2
\big(A_2B_2^\top - A_1B_1^\top\big)
\]
is attained at
\[
\begin{aligned}
&A_2 = H_{\hat z}^{-1/2}\,U'_r\,\big(\Sigma'_r\big)^{1/2} , \\
&B_2 = H_{\hat x}^{-1/2}\,V'_r\,\big(\Sigma'_r\big)^{1/2} .
\end{aligned}
\]

\paragraph{Proof.}
\emph{Step 1 (complete the square).}
Recall the definitions used in the main text,
\[
\begin{aligned}
\Delta W_2 &= A_2B_2^\top-A_1B_1^\top, \\
T &= A_1B_1^\top-H_{\hat z}^{-1}gH_{\hat x}^{-1}, \\
R &= A_2B_2^\top-T
   = \Delta W_2+H_{\hat z}^{-1}gH_{\hat x}^{-1}.
\end{aligned}
\]
The quadratic model derived above is
\[
\begin{aligned}
\Delta\mathcal L_2(\Delta W_2)
&\overset{\mathrm{K\text{-}FAC}}{\approx} L_2(\Delta W_2) \\
&=\mathrm{tr}(g^\top\Delta W_2) \\
&\quad+\frac12\,\mathrm{tr}\big(
\Delta W_2^\top H_{\hat z}\Delta W_2H_{\hat x}\big).
\end{aligned}
\]
From the definition of $R$,
$\Delta W_2=R-H_{\hat z}^{-1}gH_{\hat x}^{-1}$. Substituting this expression
into $L_2$ and expanding gives
\[
\begin{aligned}
L_2(\Delta W_2)
={}&\mathrm{tr}(g^\top R)
-\mathrm{tr}\big(g^\top H_{\hat z}^{-1}gH_{\hat x}^{-1}\big) \\
&+\tfrac12\mathrm{tr}(R^\top H_{\hat z}RH_{\hat x})
-\mathrm{tr}(g^\top R) \\
&+\tfrac12\mathrm{tr}\big(
g^\top H_{\hat z}^{-1}gH_{\hat x}^{-1}\big) \\
={}&\tfrac12\mathrm{tr}(R^\top H_{\hat z}RH_{\hat x})
-\tfrac12\mathrm{tr}\big(
g^\top H_{\hat z}^{-1}gH_{\hat x}^{-1}\big) \\
={}&\tfrac12\mathrm{tr}(R^\top H_{\hat z}RH_{\hat x})+\mathrm{const.}
\end{aligned}
\]
Here the two terms linear in $R$ cancel; the expansion uses cyclicity of the
trace, symmetry of the second moments, and
$H_{\hat z}(H_{\hat z}^{-1}gH_{\hat x}^{-1})H_{\hat x}=g$.
The second term is independent of $A_2$ and $B_2$. The minimizer is therefore
obtained by minimizing the first term.

\emph{Step 2 (whiten the approximation error).}
The whitened target is
\[
\begin{aligned}
H_{\hat z}^{1/2}TH_{\hat x}^{1/2}
={}&H_{\hat z}^{1/2}A_1B_1^\top H_{\hat x}^{1/2} \\
&-H_{\hat z}^{-1/2}gH_{\hat x}^{-1/2}
=Y_0-M.
\end{aligned}
\]
Consequently,
\[
\begin{aligned}
H_{\hat z}^{1/2}RH_{\hat x}^{1/2}
={}&H_{\hat z}^{1/2}A_2B_2^\top H_{\hat x}^{1/2} \\
&-(Y_0-M),
\end{aligned}
\]
and, using the symmetry of the square roots,
\[
\begin{aligned}
\mathrm{tr}(R^\top H_{\hat z}RH_{\hat x})
&=\big\lVert H_{\hat z}^{1/2}RH_{\hat x}^{1/2}\big\rVert_F^2 \\
&=\big\lVert
H_{\hat z}^{1/2}A_2B_2^\top H_{\hat x}^{1/2}-(Y_0-M)
\big\rVert_F^2.
\end{aligned}
\]
This target has a direct natural-gradient interpretation. Since
$D_{\mathrm{NG}}=-H_{\hat z}^{-1}gH_{\hat x}^{-1}$,
\[
\begin{aligned}
T&=A_1B_1^\top+D_{\mathrm{NG}}, \\
Y_0-M
&=H_{\hat z}^{1/2}
\big(A_1B_1^\top+D_{\mathrm{NG}}\big)H_{\hat x}^{1/2}.
\end{aligned}
\]
Thus $Y_0-M$ is the whitened adapter obtained by adding the unconstrained
K-FAC natural-gradient step \citep{amari1998natural} to the Stage-1 adapter.

\emph{Step 3 (transfer the rank constraint).}
Combining Steps~1 and 2, the K-FAC quadratic model is, up to an additive
constant independent of $A_2$ and $B_2$,
\[
\begin{aligned}
\Delta\mathcal L_2(\Delta W_2)
&\overset{\mathrm{K\text{-}FAC}}{\approx} L_2(\Delta W_2) \\
&=\tfrac12\mathrm{tr}(R^\top H_{\hat z}RH_{\hat x})+\mathrm{const.} \\
&=\tfrac12\big\lVert
H_{\hat z}^{1/2}A_2B_2^\top H_{\hat x}^{1/2}-(Y_0-M)
\big\rVert_F^2+\mathrm{const.}
\end{aligned}
\]
The additive constant and the positive factor $1/2$ do not affect the
minimizer, so they may be dropped in the rank-constrained problem below.
The matrices expressible as $A_2B_2^\top$, with
$A_2\in\mathbb{R}^{m\times r}$ and $B_2\in\mathbb{R}^{n\times r}$, are exactly
the matrices of rank at most $r$. Since multiplication by the invertible square
roots preserves rank (Lemma~2), the problem becomes
\[
\min_{\operatorname{rank}(A_2B_2^\top)\le r}
\big\lVert
H_{\hat z}^{1/2}A_2B_2^\top H_{\hat x}^{1/2}-(Y_0-M)
\big\rVert_F^2.
\]

\emph{Step 4 (truncate and transform back).}
By Lemma~1 (Eckart--Young), a minimizer is the rank-$r$ truncated SVD:
\[
\begin{aligned}
H_{\hat z}^{1/2}A_2B_2^\top H_{\hat x}^{1/2}
&=\operatorname{SVD}_r(Y_0-M) \\
&=U'_r\Sigma'_r(V'_r)^\top.
\end{aligned}
\]
Multiplying by the inverse square roots and splitting the retained singular
values symmetrically then gives
\[
\begin{aligned}
A_2B_2^\top
={}&H_{\hat z}^{-1/2}
U'_r\Sigma'_r(V'_r)^\top
H_{\hat x}^{-1/2} \\
={}&\big(H_{\hat z}^{-1/2}U'_r(\Sigma'_r)^{1/2}\big) \\
&\quad\cdot
\big(H_{\hat x}^{-1/2}V'_r(\Sigma'_r)^{1/2}\big)^\top .
\end{aligned}
\]
Thus the factors are precisely the $A_2$ and $B_2$ stated above.
$\blacksquare$

\paragraph{Remark (why the constraint is placed on the sum).}
An alternative is to constrain the increment,
$\mathrm{rank}(\Delta W_2)\le r$, solve for $\Delta W_2$, and then project the
sum back onto the budget. That is not equivalent. The constraint deployment
imposes is on the stored adapter, and $\mathrm{rank}(\Delta W_2)\le r$ is
neither sufficient for it---the sum of two rank-$r$ matrices reaches rank
$2r$---nor necessary, since a full-rank $\Delta W_2$ can leave the sum at rank
$r$. Constraining the increment therefore requires a projection outside the
optimization, after which the result is no longer the minimizer of any stated
objective and no descent guarantee follows. Constraining the sum avoids both.

\section{Damping and Constants}
\label{app:damping}

Numerical damping is applied only to the two second-moment factors: the input
factor $H_{\hat x}$ and the output factor $H_{\hat z_0}$ in Stage~1 or
$H_{\hat z}$ in Stage~2. It is applied immediately before constructing their
matrix square roots or inverse square roots. Neither the gradient $g$ nor the
cross moment $C$ is damped. We use $\varepsilon_i$ for the input factor and
$\varepsilon_o$ for the output factor.

\paragraph{Stage 1.}
Before eigendecomposition, Stage~1 damps each second-moment factor by adding a
ridge proportional to its mean diagonal:
\[
\begin{aligned}
\widetilde H_{\hat x}
&=H_{\hat x}
+\varepsilon_i\frac{\operatorname{tr}(H_{\hat x})}{n}I_n, \\
\widetilde H_{\hat z_0}
&=H_{\hat z_0}
+\varepsilon_o\frac{\operatorname{tr}(H_{\hat z_0})}{m}I_m.
\end{aligned}
\]
We use $\varepsilon_i=3\times10^{-2}$ and
$\varepsilon_o=10^{-2}$. The regularized input factor is used to construct
$H_{\hat x}^{-1/2}$, whereas the regularized output factor is used to construct
both $H_{\hat z_0}^{1/2}$ and $H_{\hat z_0}^{-1/2}$.

\paragraph{Stage 2.}
Unlike Stage~1, Stage~2 applies damping after eigendecomposition. For each
second-moment factor, negative eigenvalues are first clamped to zero, after
which the corresponding damping factor times the largest eigenvalue is added
uniformly to the spectrum. We use $\varepsilon_i=10^{-2}$ for $H_{\hat x}$ and
$\varepsilon_o=10^{-2}$ for $H_{\hat z}$. The damped factors are used to
construct both $H_{\hat x}^{\pm1/2}$ and $H_{\hat z}^{\pm1/2}$.

\paragraph{Constants.}
All methods use rank $r=64$ for every compensated linear layer.
Table~\ref{tab:hparam} reports the Stage-1 shrinkage factor, Stage-2 relative
step size, and Stage-1 calibration size used for each model and bit-width.
Stage~2 uses $16\times512$ calibration tokens in all settings.

\begin{center}
\small
\setlength{\tabcolsep}{3pt}
\begin{tabular}{@{}l c c c c@{}}
\toprule
Model & Bits & Stage-1 $\eta$ & Stage-2 $\tau$ & S1 calib. \\
\midrule
Qwen3-4B     & $2$       & $0.1$         & $0.2$         & $128\times2048$ \\
Qwen3-8B     & $2/3/4$   & $0.1/0.1/0.1$ & $0.2/0.3/0.5$ & $128\times2048$ \\
LLaMA-3.2-1B & $2$       & $0.1$         & $0.2$         & $128\times2048$ \\
\bottomrule
\end{tabular}
\captionof{table}{Hyperparameters and Stage-1 calibration sizes for all QuIP\#
experiments. The shrinkage factor is fixed to $\eta=0.1$ across these models and
bit-widths; the Stage-2 relative step size $\tau$ is selected by model and
bit-width. The 3-bit RTN experiment on LLaMA-3.1-8B uses $\eta=0.3$ and $\tau=0.1$.}
\label{tab:hparam}
\end{center}

\section{Algorithm}
\label{app:algorithm}

Algorithm~\ref{alg:ours} summarizes the complete two-stage procedure of
Sec.~\ref{sec:method}: the Stage-1 sweep that solves the Fisher-weighted
asymmetric objective for the target linear layers in each block, followed by
the Stage-2 sweep that applies the rank-constrained natural-gradient correction to the same layers.

\begin{algorithm}[tb]
\caption{Two-stage closed-form low-rank compensation}
\label{alg:ours}
\textbf{Input}: original model $\mathcal{M}$, quantized model
$\hat{\mathcal{M}}$, Stage-1 and Stage-2 calibration sets
$\mathcal{D}_1,\mathcal{D}_2$, rank $r$, shrinkage $\eta$, relative step
$\tau$\\
\textbf{Output}: final factors $A_2, B_2$ for every linear layer\\
\hspace*{1.1em}($A \in \mathbb{R}^{m\times r}$, $B \in \mathbb{R}^{n\times r}$,
rank $r$)\\
\textbf{Convention}: all collected statistics are averages over calibration
tokens
\begin{algorithmic}[1]
\STATE \textbf{Stage 1} \COMMENT{Theorem 1}
\STATE forward and backward on $\hat{\mathcal{M}}$ using $\mathcal{D}_1$
$\to H_{\hat z_0}$
\COMMENT{once, without adapters}
\FOR{block $b = 0, \dots, N_b{-}1$}
\STATE forward on $\mathcal{M}, \hat{\mathcal{M}}$ using $\mathcal{D}_1$
$\to H_{\hat x}, \tilde H_x$
\FOR{each of the $N_m$ linear layers in block $b$}
\STATE $C \gets W\tilde H_x - \hat W H_{\hat x}$
\STATE $U\Sigma V^\top \gets \mathrm{SVD}_r(H_{\hat z_0}^{1/2} C H_{\hat x}^{-1/2})$
\STATE $A_1 \gets \eta\, H_{\hat z_0}^{-1/2} U \Sigma^{1/2}$
\STATE $B_1 \gets H_{\hat x}^{-1/2} V \Sigma^{1/2}$
\STATE attach $A_1B_1^\top$ to the layer
\ENDFOR
\ENDFOR
\STATE \textbf{Stage 2} \COMMENT{Theorem 2}
\FOR{block $b = 0, \dots, N_b{-}1$}
\STATE forward and backward on $\hat{\mathcal{M}}$ using $\mathcal{D}_2$
$\to H_{\hat x}, H_{\hat z}, g$
\FOR{each of the $N_m$ linear layers in block $b$}
\STATE $Y_0 \gets H_{\hat z}^{1/2} A_1B_1^\top H_{\hat x}^{1/2}$
\STATE $M \gets H_{\hat z}^{-1/2} g H_{\hat x}^{-1/2}$
\STATE $\alpha \gets \min(1, \tau \lVert Y_0\rVert_F / \lVert M\rVert_F)$
\STATE $U'\Sigma'V'^\top \gets \mathrm{SVD}(Y_0 - \alpha M)$
\STATE $A_2 \gets H_{\hat z}^{-1/2} U'_r (\Sigma'_r)^{1/2}$
\STATE $B_2 \gets H_{\hat x}^{-1/2} V'_r (\Sigma'_r)^{1/2}$
\STATE $\Delta W_2 \gets A_2B_2^\top-A_1B_1^\top$
\STATE replace the layer's adapter with $A_2B_2^\top$
\ENDFOR
\ENDFOR
\STATE \textbf{return} $A_2, B_2$ for every linear layer
\end{algorithmic}
\end{algorithm}

\section{Diagnostics for the Two Stages}
\label{app:diagnostics}

This appendix reports the two diagnostics referenced in
Sec.~\ref{sec:method}: the out-of-sample stability of the Stage-1 singular
values, which motivates the shrinkage factor $\eta$, and the sweep of the
Stage-2 relative step size $\tau$, which motivates the trust-ratio scaling
and its clip at one.

\subsection{Stage-1 Shrinkage Diagnostic}
\label{app:svshrink}

\begin{figure}[!t]
    \centering
    \includegraphics[width=\columnwidth]{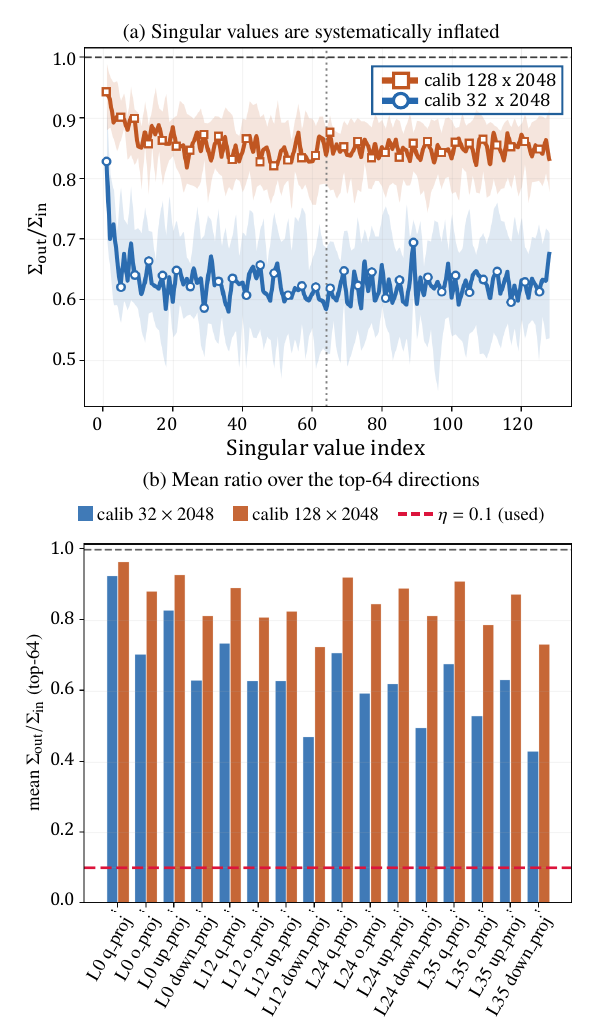}
    \caption{Out-of-sample stability of the singular values used by Stage~1 on
    Qwen3-4B quantized to $2$ bits.
    (a) Held-out-to-in-sample ratio $\Sigma_{\mathrm{out}}/\Sigma_{\mathrm{in}}$
    as a function of the singular-value index for the first $128$ directions.
    (b) Its mean over the retained top-$64$ directions for each sampled linear
    layer; the dashed line marks the shrinkage factor $\eta=0.1$ used in all
    QuIP\# experiments.
    Quadrupling the calibration set moves every linear layer toward one,
    indicating that finite-sample estimation inflates the fitted spectrum and
    motivating the shrinkage factor $\eta$.}
    \label{fig:sv-shrink}
\end{figure}

For calibration sizes $32\times2048=65{,}536$ and
$128\times2048=262{,}144$ WikiText-2 training tokens, we divide the calibration
sequences evenly into disjoint subsets $S_{\mathrm{in}}$ and
$S_{\mathrm{out}}$. Each subset contains $16$ sequences in the
$32$-sequence setting and $64$ sequences in the $128$-sequence setting. We
define the split-specific cross moments as
\[
\begin{aligned}
C_{\mathrm{in}}
&=\Ex_{t\in S_{\mathrm{in}}}[e_t\hat x_t^\top], \\
C_{\mathrm{out}}
&=\Ex_{t\in S_{\mathrm{out}}}[e_t\hat x_t^\top].
\end{aligned}
\]
Thus, $C_{\mathrm{in}}$ and $C_{\mathrm{out}}$ are computed from non-overlapping
halves containing $16\times2048$ tokens each in the smaller setting and
$64\times2048$ tokens each in the larger setting.
Using the shared $H_{\hat z_0}$ and $H_{\hat x}$, we compute
\[
\begin{aligned}
U\Sigma_{\mathrm{in}}V^\top
&=\operatorname{SVD}_{128}\!\left(
H_{\hat z_0}^{1/2}C_{\mathrm{in}}H_{\hat x}^{-1/2}
\right), \\
U\Sigma_{\mathrm{out}}V^\top
&=\operatorname{SVD}_{128}\!\left(
H_{\hat z_0}^{1/2}C_{\mathrm{out}}H_{\hat x}^{-1/2}
\right).
\end{aligned}
\]
Here, the diagonal entries of $\Sigma_{\mathrm{in}}$ and
$\Sigma_{\mathrm{out}}$ are the in-sample and held-out singular values,
respectively. A ratio below one indicates that selection on
$S_{\mathrm{in}}$ inflated the fitted singular value.

Fig.~\ref{fig:sv-shrink}(a) reports the first $128$ directions from $16$
sampled linear layers: layers $0$, $12$, $24$, and $35$, each with
\texttt{q\_proj}, \texttt{o\_proj}, \texttt{up\_proj}, and
\texttt{down\_proj}. Panel (b) and all summary statistics use only the retained
top-$64$ directions, matching the adapter rank. Only $0.68\%$ of these ratios
exceed one. Increasing the calibration data fourfold raises their mean from
$0.6403$ to $0.8517$, with every sampled linear layer moving toward one. The value
$\eta=0.1$ used in our experiments lies below these spectral ratios, suggesting
that it controls not only finite-sample inflation but also the magnitude of the
cascaded Stage-1 updates.

\begin{center}
\small
\setlength{\tabcolsep}{8pt}
\begin{tabular}{@{}l c@{}}
\toprule
Stage-1 shrinkage $\eta$ & C4 perplexity $\downarrow$ \\
\midrule
No adapter & $29.2146$ \\
$0.1$      & $27.5103$ \\
$0.2$      & $\mathbf{27.1697}$ \\
$0.3$      & $27.6247$ \\
$0.5$      & $35.7916$ \\
\bottomrule
\end{tabular}
\captionof{table}{Stage-1-only shrinkage sweep on held-out C4 for Qwen3-4B
quantized to $2$ bits with QuIP\#. Bold denotes the lowest perplexity.}
\label{tab:stage1-eta-sweep}
\end{center}

Table~\ref{tab:stage1-eta-sweep} provides an end-to-end complement to the
spectral diagnostic. Moderate shrinkage consistently improves over the
uncompensated model, whereas the larger value $\eta=0.5$ is unstable. Although
$\eta=0.2$ gives the best C4 perplexity after Stage~1 alone, we use
$\eta=0.1$ in the complete method because $\eta$ is selected for the
two-stage pipeline rather than to optimize the intermediate Stage-1 model:
the more conservative Stage-1 adapter leaves greater scope for the subsequent
Stage-2 natural-gradient correction.
\subsection{Stage-2 Relative Step Size}
\label{app:tausweep}

\begin{figure}[t]
    \centering
    \includegraphics[width=\columnwidth,keepaspectratio]{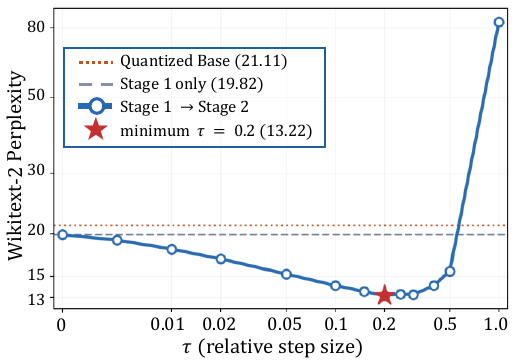}
    \caption{WikiText-2 test perplexity as a function of the Stage-2 relative
    step size $\tau$ on Qwen3-4B quantized to $2$ bits with QuIP\#. The ordinate
    is logarithmic. At $\tau=0$, Stage~2 only re-truncates the rank-$64$ Stage-1
    adapter and reproduces its perplexity. The broad minimum at
    $\tau=0.2$--$0.3$ shows the benefit of the gradient correction, while an
    excessively large step exceeds the regime in which the local quadratic
    model is accurate.}
    \label{fig:tau-sweep}
\end{figure}

Fig.~\ref{fig:tau-sweep} directly evaluates the scaling in
Eq.~\eqref{eq:stage2-svd-alpha}. We fix the rank-$64$ Stage-1 adapter obtained with
$\eta=0.1$ from $128\times2048=262{,}144$ WikiText-2 training tokens. For
each of $12$ values
\[
\begin{aligned}
\tau\in\{&0,0.005,0.01,0.02,0.05,0.1,\\
           &0.2,0.25,0.3,0.4,0.5,1.0\},
\end{aligned}
\]
we solve Stage~2 independently, collecting $g$, $H_{\hat x}$, and
$H_{\hat z}$ from $16\times512=8{,}192$ WikiText-2 training tokens, and
evaluate perplexity on the disjoint WikiText-2 test split with sequence length
$2048$.

At $\tau=0$, $\alpha=0$ and the SVD target is $Y_0$, which already has rank at
most $64$. The measured perplexity $19.8200$ therefore matches Stage~1
($19.8194$), confirming that re-truncation alone provides no gain. Turning on
the gradient term lowers perplexity to $13.22$, $13.28$, and $13.24$ at
$\tau=0.2$, $0.25$, and $0.3$, respectively, giving a broad stable region rather
than a sharp optimum. Perplexity increases rapidly for larger steps, reaching
$14.07$ at $\tau=0.4$, $15.51$ at $\tau=0.5$, and $83.38$ at $\tau=1$.
This sweep supports scaling the step relative to the current adapter and
clipping the full-step coefficient at one.

\section{Adapter Construction Cost}
\label{app:cost}

At inference, every method considered here stores one rank-$64$ adapter per
linear layer, so the additional cost of our method is confined to offline
adapter construction. All timings below are for one run on a single NVIDIA
A100 GPU with $40$\,GB of memory.

\subsection{Construction Time}

For Qwen3-8B at $2$ bits, the two stages take approximately 53m with a
precomputed $H_{\hat z_0}$ cache. Constructing this $H_{\hat z_0}$ cache
takes approximately 10 minutes, raising the first-run total to 1h~3m.
The cache is built once and can be reused while the model, calibration
data, and calibration size remain unchanged, so this overhead is a
one-time cost rather than a per-run cost. The baseline methods take
16--37 minutes for Qwen3-8B.

\begin{center}
\small
\setlength{\tabcolsep}{7pt}
\begin{tabular}{@{}l cccc@{}}
\toprule
Bit width & EoRA & QERA & ProjQ & LQ-LoRA \\
\midrule
$2$ bit & 22m & 21m & 16m & 37m \\
\bottomrule
\end{tabular}
\captionof{table}{Measured Qwen3-8B 2-bit baseline construction time, rounded to the
nearest minute. Each run uses rank $64$ and $128\times2048$ WikiText-2
calibration tokens.}
\label{tab:cost-time}
\end{center}

\begin{center}
\small
\setlength{\tabcolsep}{7pt}
\begin{tabular}{@{}c cc@{}}
\toprule
Bit width & Stage~1 (cached) & Stage~2 \\
\midrule
$2$ bit & 35m & 18m \\
\bottomrule
\end{tabular}
\captionof{table}{Qwen3-8B construction time 2-bit. Stage~1
assumes a precomputed $H_{\hat z_0}$ cache. Stage~1 uses $128\times2048$
calibration tokens, whereas Stage~2 uses $16\times512$ tokens; both use rank
$64$.}
\label{tab:cost-ours-time}
\end{center}

\subsection{Storage and Memory}

\begin{center}
\small
\setlength{\tabcolsep}{6pt}
\begin{tabular}{@{}l r@{}}
\toprule
Artifact & Size \\
\midrule
Stage~1 adapter ($\eta=0.1$, rank $64$, FP16) & 349\,MB \\
Stage~2 adapter ($\tau=0.2$, rank $64$, FP16) & 349\,MB \\
\bottomrule
\end{tabular}
\captionof{table}{Qwen3-8B adapter sizes after Stage~1 and Stage~2 at $2$ bits.}
\label{tab:cost-artifacts}
\end{center}

\begin{center}
\small
\setlength{\tabcolsep}{5pt}
\begin{tabular}{@{}l r@{}}
\toprule
Qwen3-8B component & Peak GPU \\
\midrule
$H_{\hat z_0}$ cache construction & 30.2\,GiB \\
Stage~1, cache hit & $\approx 33$\,GB \\
Stage~1, FP offload & 31.8\,GB \\
Stage~2 & 28--31\,GB \\
\bottomrule
\end{tabular}
\captionof{table}{Observed memory use on one NVIDIA A100 GPU with
$40$\,GB of device memory.}
\label{tab:cost-mem}
\end{center}

The rank-$64$ adapters occupy 264--349\,MB. The reusable output-statistic cache
is substantially larger, at 29.7\,GiB for Qwen3-4B and 47.5\,GiB for Qwen3-8B.
Peak Qwen3-8B device memory remains within the 40\,GB capacity of the test GPU.

\section{Computation Count}
\begin{table*}[t]
\centering
\scriptsize
\setlength{\tabcolsep}{2.5pt}
\resizebox{0.85\textwidth}{!}{%
\begin{tabular}{@{}l rr rrrrrrr r@{}}
\toprule
Method & WT2 $\downarrow$ & C4 $\downarrow$ & ARC-C & ARC-E & BoolQ
& HellaS & OBQA & PIQA & WinoG & CSQA Avg $\uparrow$ \\
\midrule
\multicolumn{11}{@{}l}{\textit{Full precision}} \\
FP16 & 9.7276 & 15.2365 & 56.66 & 80.85 & 86.67 & 75.04 & 41.20
& 77.53 & 67.72 & 69.38 \\
\midrule
\multicolumn{11}{@{}l}{\textit{4-bit QuIP\#}} \\
QuIP\# & 9.9923 & 15.5443 & 55.46 & 79.38 & 86.64 & 74.10 & 40.40
& 77.64 & 68.19 & 68.83 \\
+ EoRA & 9.9443 & 15.5276 & 55.20 & 79.55 & 86.57 & 74.10 & 40.00
& 77.64 & 68.27 & 68.76 \\
+ QERA & 9.9455 & 15.5286 & 54.86 & 79.55 & \textbf{86.70} & 74.18
& 40.80 & \textbf{77.86} & 68.43 & 68.91 \\
+ ProjQ & 9.9639 & 15.5482 & 55.20 & 78.87 & 86.48 & 74.17 & 40.40
& 77.31 & 69.46 & 68.84 \\
+ LQ-LoRA & 9.8663 & 15.4630 & \textbf{56.06} & \textbf{79.71}
& \textbf{86.70} & 74.18 & 40.60 & 77.31 & 68.43 & 69.00 \\
+ Ours & \textbf{8.2172} & \textbf{14.1476} & 55.63 & 79.17 & 86.36
& \textbf{75.84} & \textbf{41.40} & 77.64 & \textbf{69.85}
& \textbf{69.41} \\
\midrule
\multicolumn{11}{@{}l}{\textit{3-bit QuIP\#}} \\
QuIP\# & 10.4808 & 16.2979 & 53.33 & 78.70 & 86.12 & 72.00 & 42.40
& 76.01 & 68.11 & 68.10 \\
+ EoRA & 10.3730 & 16.2178 & 54.18 & \textbf{79.46} & 86.02 & 72.11
& 42.60 & 76.28 & 69.14 & 68.54 \\
+ QERA & 10.3764 & 16.2117 & 53.92 & 79.08 & 85.96 & 72.08 & 42.60
& 76.33 & 69.85 & 68.55 \\
+ ProjQ & 10.3624 & 16.2127 & \textbf{54.52} & 79.38 & \textbf{86.18}
& 72.02 & 42.20 & 76.22 & 68.90 & 68.49 \\
+ LQ-LoRA & 10.2391 & 15.9019 & 53.41 & 78.75 & 85.93 & 72.50
& 41.80 & 76.77 & 68.35 & 68.22 \\
+ Ours & \textbf{8.5375} & \textbf{14.5620} & 53.84 & 79.25 & 86.09
& \textbf{74.12} & \textbf{43.20} & \textbf{77.15} & \textbf{70.09}
& \textbf{69.11} \\
\midrule
\multicolumn{11}{@{}l}{\textit{2-bit QuIP\#}} \\
QuIP\# & 12.4285 & 19.1851 & 45.39 & 69.23 & 83.15 & 65.16 & 35.60
& 74.43 & 66.69 & 62.81 \\
+ EoRA & 12.1971 & 18.9834 & 45.56 & 70.75 & 83.00 & 65.51 & 35.80
& 74.43 & 66.14 & 63.03 \\
+ QERA & 12.2233 & 18.9668 & 45.31 & 70.96 & 83.00 & 65.60 & 36.00
& 74.65 & 65.90 & 63.06 \\
+ ProjQ & 12.1526 & 18.9088 & 46.42 & 70.75 & 82.81 & 65.31 & 35.20
& 74.32 & 66.69 & 63.07 \\
+ LQ-LoRA & 11.6161 & 17.9490 & 47.44 & \textbf{74.54}
& \textbf{83.76} & 66.55 & 36.20 & 74.21 & 67.32 & 64.29 \\
+ Ours & \textbf{10.2592} & \textbf{17.1713} & \textbf{47.53} & 71.51
& 83.61 & \textbf{67.96} & \textbf{37.20} & \textbf{75.08}
& \textbf{68.67} & \textbf{64.51} \\
\bottomrule
\end{tabular}%
}
\caption{Full Qwen3-8B results under QuIP\# quantization. WT2 and C4 report
perplexity; the seven task columns report zero-shot accuracy, and CSQA Avg is
their unweighted mean. Best results within each bit-width are in \textbf{bold}.}
\label{tab:qwen3-8b-full}
\end{table*}

\begin{table*}[t]
\centering
\scriptsize
\setlength{\tabcolsep}{2.5pt}
\resizebox{0.85\textwidth}{!}{%
\begin{tabular}{@{}l rr rrrrrrr r@{}}
\toprule
Method & WT2 $\downarrow$ & C4 $\downarrow$ & ARC-C & ARC-E & BoolQ
& HellaS & OBQA & PIQA & WinoG & CSQA Avg $\uparrow$ \\
\midrule
\multicolumn{11}{@{}l}{\textit{2-bit QuIP\#}} \\
QuIP\# & 21.1093 & 29.2146 & 39.59 & 63.38 & 77.98 & 57.10 & 35.60
& 68.44 & 60.22 & 57.47 \\
+ EoRA & 20.9228 & 27.7091 & 41.38 & 65.66 & 76.82 & 57.64
& \textbf{38.20} & 70.67 & 60.46 & 58.69 \\
+ QERA & 20.9066 & 27.6689 & 41.30 & 65.24 & 76.91 & 57.75 & 37.60
& 70.51 & 61.09 & 58.63 \\
+ ProjQ & 21.1621 & 27.8139 & 40.70 & 65.15 & 76.97 & 57.77 & 37.00
& 70.40 & 60.85 & 58.41 \\
+ LQ-LoRA & 16.7533 & 23.3417 & 39.93 & 68.01 & \textbf{81.68}
& 59.93 & 36.20 & 72.03 & 59.83 & 59.66 \\
+ Ours & \textbf{13.2193} & \textbf{21.3614} & \textbf{41.89}
& \textbf{68.06} & 77.43 & \textbf{61.03} & 37.40 & \textbf{72.09}
& \textbf{62.43} & \textbf{60.05} \\
\bottomrule
\end{tabular}%
}
\caption{Full Qwen3-4B results at $2$ bits under QuIP\# quantization. WT2 and
C4 report perplexity; the seven task columns report zero-shot accuracy, and
CSQA Avg is their unweighted mean. Best results are in \textbf{bold}.}
\label{tab:qwen3-4b-full}
\end{table*}

\begin{table*}[t]
\centering
\scriptsize
\setlength{\tabcolsep}{2.5pt}
\resizebox{0.7\textwidth}{!}{%
\begin{tabular}{@{}l rrrrrrr r@{}}
\toprule
Method & ARC-C & ARC-E & BoolQ & HellaS & OBQA & PIQA & WinoG
& CSQA Avg $\uparrow$ \\
\midrule
FP16 & 36.77 & 61.91 & 63.91 & 64.18 & 37.60 & 74.92 & 60.77 & 57.15 \\
\midrule
QuIP\# & 27.73 & 43.73 & 52.02 & 46.00 & 30.40 & 64.58 & 55.17 & 45.66 \\
+ EoRA & 27.47 & 46.09 & 52.20 & 47.59 & 31.80 & 66.00 & 53.83 & 46.43 \\
+ QERA & 27.30 & 46.00 & 52.17 & 47.65 & 32.00 & 65.89 & 53.04 & 46.29 \\
+ ProjQ & 28.58 & 46.13 & 51.77 & 47.30 & 31.20 & 65.89 & 54.14 & 46.43 \\
+ LQ-LoRA & \textbf{29.10} & \textbf{48.15} & 53.61 & 49.09
& 31.00 & \textbf{67.30} & 55.80 & 47.72 \\
+ Ours & 28.92 & 47.60 & \textbf{54.83} & \textbf{49.35}
& \textbf{33.20} & 67.14 & \textbf{56.67} & \textbf{48.25} \\
\bottomrule
\end{tabular}
}
\caption{Full zero-shot results for LLaMA-3.2-1B at $2$ bits under QuIP\#
quantization. Stage~1 uses $128\times2048$ WikiText-2 calibration tokens with
$\eta=0.1$, and Stage~2 uses $16\times512$ tokens with $\tau=0.2$. CSQA Avg is
the unweighted mean of the seven tasks. Best quantized results are in
\textbf{bold}; FP16 is shown only as a reference.}
\label{tab:llama32-1b-full}
\end{table*}

One pass processes the entire calibration set. Let $N_b$ be the number of
decoder blocks and $N_m$ the number of target linear layers per block
($N_m=7$ for the Qwen3 models). A forward pass denotes a standard forward
evaluation, and a backward pass denotes the corresponding gradient computation.
Stage~1 first performs one forward and one backward pass on the quantized model
without adapters to collect $H_{\hat z_0}$. It then sweeps the blocks from front
to back. At each block, the same calibration tokens are passed once through the
full-precision model and once through the progressively compensated quantized
model, yielding $2N_b$ forward passes that provide $H_{\hat x}$ and
$\tilde H_x$. Together with the initial quantized-only forward pass, Stage~1
therefore uses $2N_b+1$ forward passes and one backward pass. Stage~2 operates
on a single quantized model with the Stage-1 adapters already installed and
performs one forward and one backward pass together at each block, jointly
collecting $g$, $H_{\hat x}$, and $H_{\hat z}$ for all $N_m$ target linear layers. It
therefore uses $N_b$ forward and $N_b$ backward passes.
\begin{center}
\scriptsize
\setlength{\tabcolsep}{3pt}
\resizebox{\columnwidth}{!}{%
\begin{tabular}{@{}l ccc ccc@{}}
\toprule
& \multicolumn{3}{c}{General} & \multicolumn{3}{c}{Qwen3-4B ($N_b=36$)} \\
\cmidrule(lr){2-4}\cmidrule(l){5-7}
Stage & Forward & Backward & SVD & Forward & Backward & SVD \\
\midrule
Stage~1 & $2N_b+1$ & $1$ & $N_bN_m$ & $73$ & $1$ & $252$ \\
Stage~2 & $N_b$ & $N_b$ & $N_bN_m$ & $36$ & $36$ & $252$ \\
Total   & $3N_b+1$ & $N_b+1$ & $2N_bN_m$ & $109$ & $37$ & $504$ \\
\bottomrule
\end{tabular}%
}
\captionof{table}{Pass and SVD counts for constructing one rank-$64$ adapter
set. Each block-level backward pass jointly supplies statistics for all $N_m$
target linear layers.}
\label{tab:computation-count}
\end{center}

The pass sizes differ substantially: Stage~1 uses $128\times2048=262{,}144$
tokens per pass, whereas Stage~2 uses $16\times512=8{,}192$, which is $32$
times smaller. For Qwen3-4B this corresponds to $19{,}136{,}512$ versus
$294{,}912$ forward tokens, a factor of approximately $65$, while the backward
volumes are similar ($262{,}144$ versus $294{,}912$ tokens). Thus the paired
full-precision and quantized forwards in Stage~1 dominate the total computation,
despite Stage~2 performing a backward pass at every block. When a compatible
$H_{\hat z_0}$ cache is reused, the single Stage-1 backward pass is omitted.

\section{Full Per-Model and Per-Task Results}
\label{app:full-results}

The following tables provide the complete breakdown behind the aggregate
results in Sec.~\ref{sec:experiments}. We report all three bit-widths for
Qwen3-8B and the $2$-bit setting for Qwen3-4B and LLaMA-3.2-1B, including the
seven individual zero-shot tasks used to compute CSQA Avg. For the Qwen3
models, we additionally report perplexity on WikiText-2 and the held-out C4
corpus.

\end{document}